\def\year{2022}\relax
\documentclass[letterpaper]{article} % DO NOT CHANGE THIS
\usepackage{aaai22}  % DO NOT CHANGE THIS
\usepackage{times}  % DO NOT CHANGE THIS
\usepackage{helvet}  % DO NOT CHANGE THIS
\usepackage{courier}  % DO NOT CHANGE THIS
\usepackage[hyphens]{url}  % DO NOT CHANGE THIS
\usepackage{graphicx} % DO NOT CHANGE THIS
\usepackage{natbib}  % DO NOT CHANGE THIS AND DO NOT ADD ANY OPTIONS TO IT
\usepackage{caption} % DO NOT CHANGE THIS AND DO NOT ADD ANY OPTIONS TO IT
\DeclareCaptionStyle{ruled}{labelfont=normalfont,labelsep=colon,strut=off} % DO NOT CHANGE THIS
\usepackage{algorithm}
\usepackage{algorithmic}

\usepackage{newfloat}
\usepackage{listings}
\floatstyle{ruled}
\newfloat{listing}{tb}{lst}{}
\floatname{listing}{Listing}

\usepackage{color}
\usepackage{subfigure}
\usepackage{amsmath}
\usepackage{amssymb}
\usepackage{multirow}
\usepackage{booktabs}
\usepackage{pifont}% http://ctan.org/pkg/pifont
\newcommand{\cmark}{\ding{51}}%
\usepackage{makecell}
\usepackage[pagebackref=true,breaklinks=true,colorlinks,bookmarks=false]{hyperref}

\title{Homography Decomposition Networks for Planar Object Tracking }

\title{Homography Decomposition Networks for Planar Object Tracking}
\author {
	Xinrui Zhan\textsuperscript{\rm 1},
	Yueran Liu\textsuperscript{\rm 1},
	Jianke Zhu\textsuperscript{\rm 1,2\thanks{corresponding authors}},
	Yang Li\textsuperscript{\rm 3\footnotemark[1]} 
}
\affiliations {
	\textsuperscript{\rm 1} Zhejiang University, Hangzhou, China\\
	\textsuperscript{\rm 2} Alibaba-Zhejiang University Joint Institute of Frontier Technologies\\
	\textsuperscript{\rm 3} East China Normal University, Shanghai, China\\
	\{xrzhan, liuyueran97, jkzhu\}@zju.edu.cn, yli@cs.ecnu.edu.cn}
\usepackage{bibentry}
\begin{document}
	
	\maketitle
	
	\begin{abstract}
		Planar object tracking plays an important role in AI applications, such as robotics, visual servoing, and visual SLAM.
		Although the previous planar trackers work well in most scenarios, it is still a challenging task due to the rapid motion and large transformation between two consecutive frames. 
		The essential reason behind this problem is that the condition number of such a non-linear system changes unstably when the searching range of the homography parameter space becomes larger.
		To this end, we propose a novel Homography Decomposition Networks~(HDN) approach that drastically reduces and stabilizes the condition number by decomposing the homography transformation into two groups.
		Specifically, a similarity transformation estimator is designed to predict the first group robustly by a deep convolution equivariant network. By taking advantage of the scale and rotation estimation with high confidence, a residual transformation is estimated by a simple regression model. Furthermore, the proposed end-to-end network is trained in a semi-supervised fashion. Extensive experiments show that our proposed approach outperforms the state-of-the-art planar tracking methods at a large margin on the challenging POT, UCSB and POIC datasets. Codes and models are available at \href{https://github.com/zhanxinrui/HDN}{https://github.com/zhanxinrui/HDN}. 
	\end{abstract}

	\section{Introduction}
	Planar object tracking is a fundamental problem in many AI applications, which aims at estimating the transformation of a planar object in videos. A reliable planar tracker is usually employed as a reasonable surrogate for 3D structure tracking methods, such as augmented reality, visual servoing, and robotics.
	Despite the encouraging progress~\cite{GOP-ESM, Gracker, LISRD, GIFT} has been made in past decades, tracking planar object robustly is still a challenging problem due to the appearance changes and large displacements between the consecutive video frames.

	\begin{figure}[t]
		\centering 
		\includegraphics[scale=0.38]{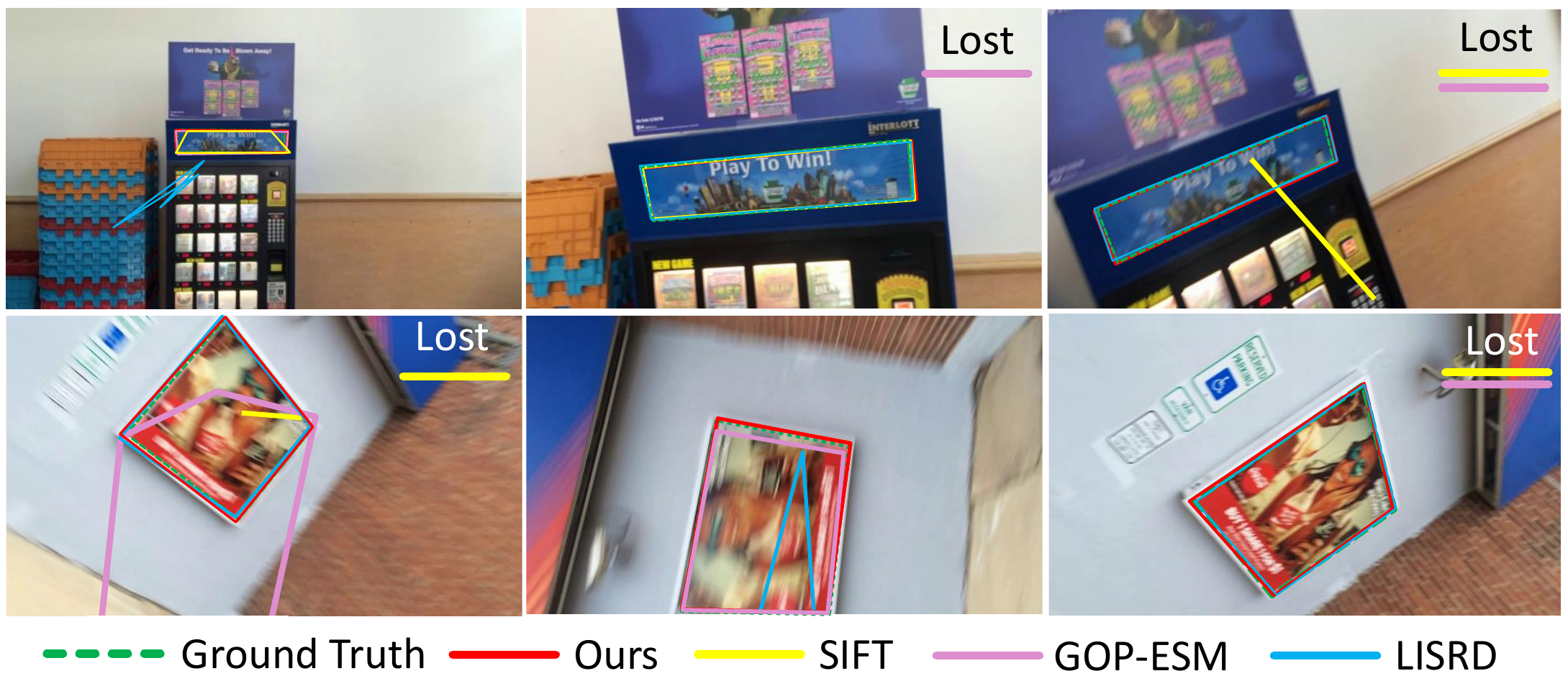}
		\caption{A comparison of our approach with state-of-the-art trackers. 
		Our method HDN obtains the robust tracking results comparing to the keypoint-based method~(SIFT, LISRD) and direct approach~(GOP-ESM). }
		\label{fig:results_plot}
	\end{figure}
	Matching the salient keypoints between two images is typically employed to recover the homography, which is widely used in planar object tracking. Since the keypoints-based method cannot make full use of the whole image, jitters commonly occur during tracking. Moreover, the conventional feature detector SIFT~\cite{SIFT} often fails in the case of large perspective transformation. The deep learning-based methods such as LISRD~\cite{LISRD} and GIFT~\cite{GIFT} introduce the invariant descriptors to tackle this problem, which still suffer from the issue of insufficient keypoints to rebuild the homography. On the other hand, the appearance-based approaches are able to take full advantage of information from the whole image. However, it is easy to get stuck in the local optima. Although the gradient-based method such as GOP-ESM \cite{GOP-ESM} can deal with the illumination changes, they are not robust to motion blurs. In addition, large displacements may lead to the fractional sampled object patches, which are harmful to the minimization-based approaches. 
	
	In contrast to the conventional approaches, the deep learning-based method becomes the promising direction for planar object tracking.~\cite{DeTone2016DeepIH, Nguyen2018UnsupervisedDH} directly regress the corners' offsets of the rectangular region, however, it cannot accommodate the large transformation. Specifically, directly estimating the homography parameterized by corner offsets with eight coefficients, the condition number of the system becomes extremely large (up to $5e^7$), which makes the system very unstable. Therefore, any perturbation may lead to tracking failure, especially with large displacement. It can be found that the condition number becomes lower with fewer parameters to be estimated. To predict four transformation parameters of an object, \cite{Black2004EigenTrackingRM, LDES} employ the rigid motion along with scaling model for the region-based trackers, which obtain the very robust tracking result. However, they do not estimate the homography transformation with eight parameters for the planar object tracking.
	
	In this paper, we propose a novel Homography Decomposition Networks approach to planar object tracking in video sequences, which decomposes the homography transformation into two groups, including a similarity group and a residual group. By estimating the similarity group firstly, the condition number of the entire system reduces substantially.
	Inspired by group convolution theory~\cite{Henriques2017WarpedCE}, we employ a rotation-scale invariant convolution operator to predict similarity robustly.
	Intuitively, this gives a very robust and good initial guess of where the object is. Then,
	the second stage predicts the residual transformation through the semi-supervised regression, where the residual transformation is the residual group with the extra error from the first stage.
	To the best of our knowledge, this is the first work to decompose homography into two stages in deep learning.

	The contribution of our work can be summarized as below: 1) a novel deep planar object tracker by decomposing the homography matrix into two groups; 2) a deep similarity equivalent estimator robustly recovers the similarity transformation; 3) an end-to-end differentiable semi-supervised model with negative samples loss bridges the gap from homography estimation; 4) experiments on challenging POT, UCSB, and POIC datasets, show that our method performs better than the state-of-the-art approaches.

	\begin{figure*}[htbp]
		\centering 
		\includegraphics[scale=0.320]{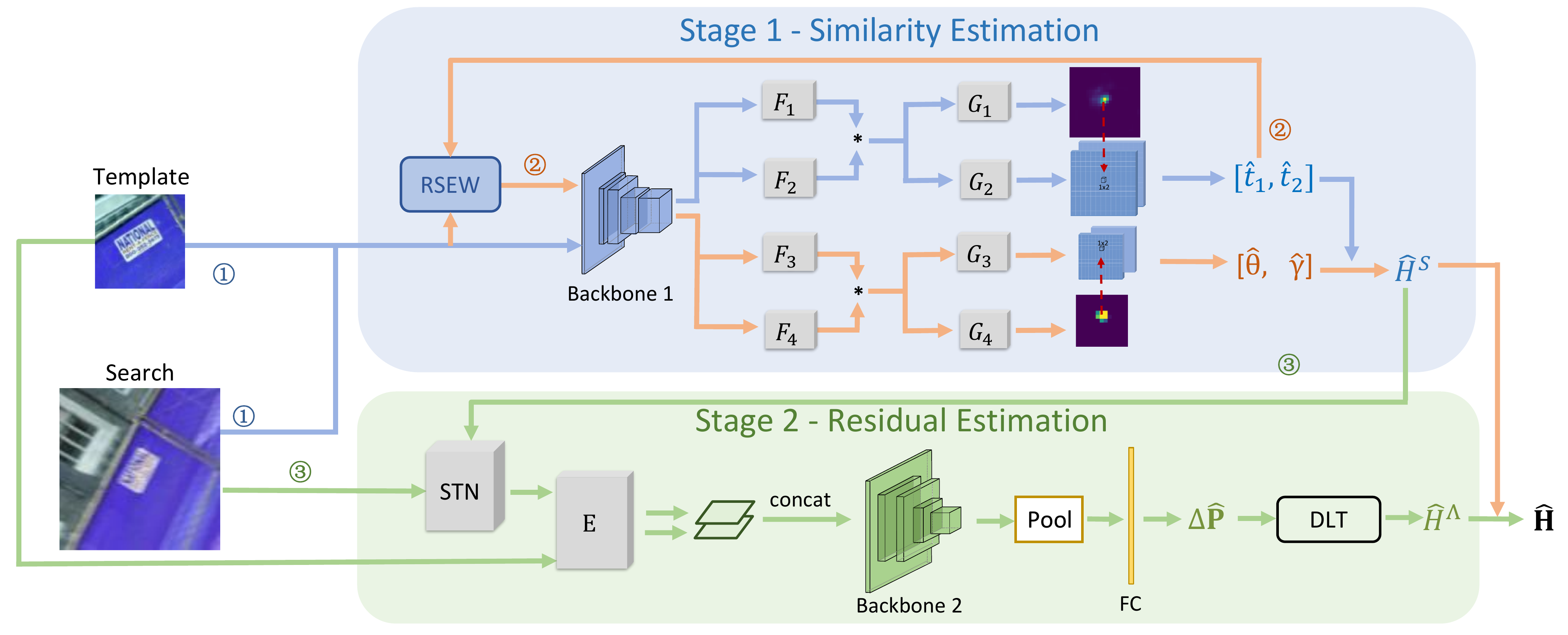}
		\caption{The tracking pipeline of HDN.  The data flows in the networks obey the order number~($1\to 2 \to 3$). 
			In Similarity Estimation, $F$ denotes the neck for feature cut and convolutional layer, and $G$ denotes the convolutional layer for generating the output map.	
			Residual Estimation Network is composed of the shared coarse feature extraction $E$, backbone, pooling layer (Pool), an FC (Fully Connected Layer) and Direct Linear Transformation (DLT)~\cite{DLT}, which converts the prediction to homography. STN~\cite{STN} is used to make the whole networks differentiable.
		}
		\label{fig:all_structure}
	\end{figure*}
	\section{Related Works}
	\subsection{Planar Object Tracking}
	The conventional planar trackers can be roughly categorized into two groups. 
	
	One category is the keypoint-based methods, which often adopt the salient feature point detection technique such as SIFT~\cite{SIFT} and match the object across images using the keypoints. 
	Then, the homography is estimated from those inlier correspondences between the reference frame and target image. Gracker~\cite{Gracker} establishes the correspondences in a geometric graph matching manner, which adopts the match-filtering and optimization strategy in order to bring robustness for more scenes and transformation. SuperGlue~\cite{SuperGlue} introduces a deep attentional GNN with a keypoint matching layer. LISRD~\cite{LISRD} presents a novel CNN-based dense descriptor, which is invariant to a group of transformations. GIFT~\cite{GIFT} uses joint learning to select the right invariance to match the descriptors. 
	
	Another group is the region-based approaches. ESM \cite{ESM} reduces the hessian matrix computation in optimization while still retaining the high convergence rates. However, SSD-based ESM is not robust to illumination changes. To tackle this issue, GOP-ESM~\cite {GOP-ESM} proposes an illumination insensitive ESM method using the gradient pyramid. Although having achieved encouraging results, these trackers usually suffer from appearance changes such as occlusions and motion blurs. UDH~\cite{Nguyen2018UnsupervisedDH} and~\cite{Tsai2019PlanarTB} approximate the corner offsets to compose the homography matrix based on the global region feature. These deep learning-based methods can only deal with the small local transformation, which are not robust to the drastic changes. Our proposed approach overcomes this issue through homography decomposition networks. Specifically, the first similarity component accommodates the large motion including scale and in-plane rotation, and the second stage predicts the small residual transformation. 
	
	\subsection{Visual Object Tracking}
	The planar tracker can be viewed as a special case of general object tracking methods. Recently, deep learning-based approaches dominate the leaderboard of mainstreamed benchmarks. SiamFC~\cite{SiamFC} first uses Siamese Network to generate two feature maps, whose correlation is further employed to locate the center of the target. For the scale changes, multi-scale pooling is adopted, which incurs the computational burden. Anchor-free-based trackers~\cite{SiamBAN, Ocean, SiamCAR} acquire the probability to be the center of the target of each point from the feature map. Another crucial problem is that the conventional CNN is not equivalent to transformation except translation, which makes it hard to predict the common transformation such as rotation. To deal with this problem, RE-SiamNets~\cite{Gupta2020RotationES} use the rotation equivalence of steerable filters. Despite the easiness to insert to other trackers, its sampling scheme is time-consuming. In this paper, we exert the correlation and the anchor-free-based method used in visual tracking to deal with the robust similarity estimation problem in HDN.

	\section{Method}
	The objective of the planar object tracking is to recover the underlying homography transformation from the template image $T$ to the $i_{th}$ input frame $I_i \in \mathbb{R}^{n\times n}$, where the size of $I_i$ is $n\times n$. 
	
	Let $\mathbf{p}=(u,v,1)^T$ be the homogenous coordinates vector of a pixel, and $I(\mathbf{p})$ denotes the pixel value of $\mathbf{p}$ in image $I$.

	Let $\mathbf{H}_{i}$ be the estimated transformation matrix (from $T$ to $I_i$). ${\mathbf{P}} = [\mathbf{p}_{lt}, \mathbf{p}_{rt}, \mathbf{p}_{rb},\mathbf{p}_{lb}]$ is the four corner points quadrilateral coordinates of an object, which is a ($3\times 4$) matrix. Specially, $ {\mathbf{P}}_{i}$ represents the coordinates of the target object in $I_i$, and ${\mathbf{P}}_{T}$ for the object coordinates in $T$. Therefore, the prediction of $ {\mathbf{P}}_{i}$ can be derived as follows:

	\begin{equation}
	{\hat{\mathbf{P}}}_{i} = \mathbf{H}_{i} \cdot {\mathbf{P}}_{T} = \prod_{k=1}^{i} \hat{\mathbf{H}}_{k} \cdot {\mathbf{P}}_{T}
	\label{eq:total_H}  
	\end{equation}
    where $\hat{\mathbf{H}}_{k}$ is the the estimated transformation from template image $T$ to the resampled image
	$\mathcal{W}(I_{k}, \mathbf{H}_{k-1}^{-1})$ for $k>1$.
	For $k=1$, $\hat{\mathbf{H}}_1$ denotes the transformation from $T$ to $I_1$.
	$\mathcal{W}(I_i, \mathbf{H})$ is a warped image for $I_i$ with respect to $\mathbf{H}$.

	Since a planar object's motions within the video sequences are continuous, it is unnecessary to predict the homography from the template $T$ to $I_i$ at every frame for a robust tracker. Therefore, a compositional matrix $\prod_{k=1}^{i} \hat{\mathbf{H}}_{k}$ is introduced to denote the cumulative transformation in the previous frames, and we only need to predict $\hat{\mathbf{H}}_{i}$ at each frame. Our deep approach adopts a forward method that only warps the input image $I_i$ to avoid the object out of view when warping the template. Moreover, this reduces the extra computational cost for extracting the deep features of the template in the backbone at each frame and stabilize the template during tracking. In our proposed HDN approach, we only compute template's deep feature only once as the template image is kept constant.
	
	\subsection{Decomposition}
	The conventional approach~\cite{DeTone2016DeepIH} directly regresses the four corners' displacements of a rectangular planar object to predict the homography. Since the homography transformation does not directly relate to the corners' movements, it is determined by the planar object's pose in 3D space. Given a vector of eight transformation parameters $\mathbf{x}=[t_1, t_2, \gamma, \theta, k_1, k_2, \nu_1, \nu_2]^T$, the homography transformation $\mathbf{H}$ can be constructed efficiently as detailed in ~\cite{Harltey2003MultipleVG}.
    To further analyse the numerical properties of the proposed approach, let $
	\Delta \mathbf{p}(\mathbf{x}) = \mathbf{H}(\mathbf{x})\cdot \mathbf{p}-\mathbf{p}$ be the difference between pixel coordinates before and after the homography transformation parameterized by $\mathbf{x}$.  
		
\begin{figure}[t]
		\centering
		\subfigure{
			\begin{minipage}[t]{0.442\linewidth}
				\centering
				\includegraphics[width=\linewidth]{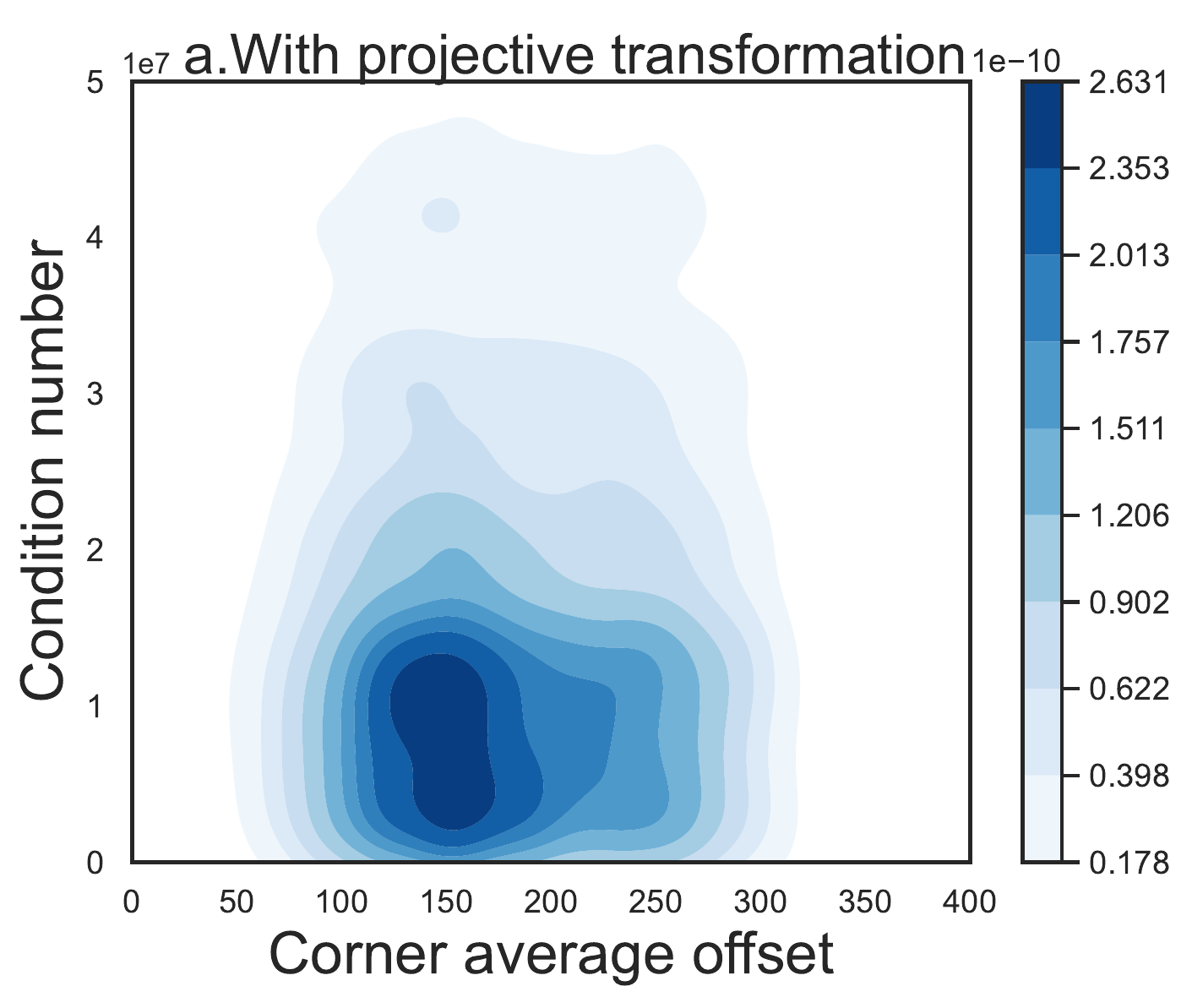}
				%\caption{fig1}
			\end{minipage}%
			\label{fig:cond_numa}
		}%
		\subfigure{
			\begin{minipage}[t]{0.442\linewidth}
				\centering
				\includegraphics[width=\linewidth]{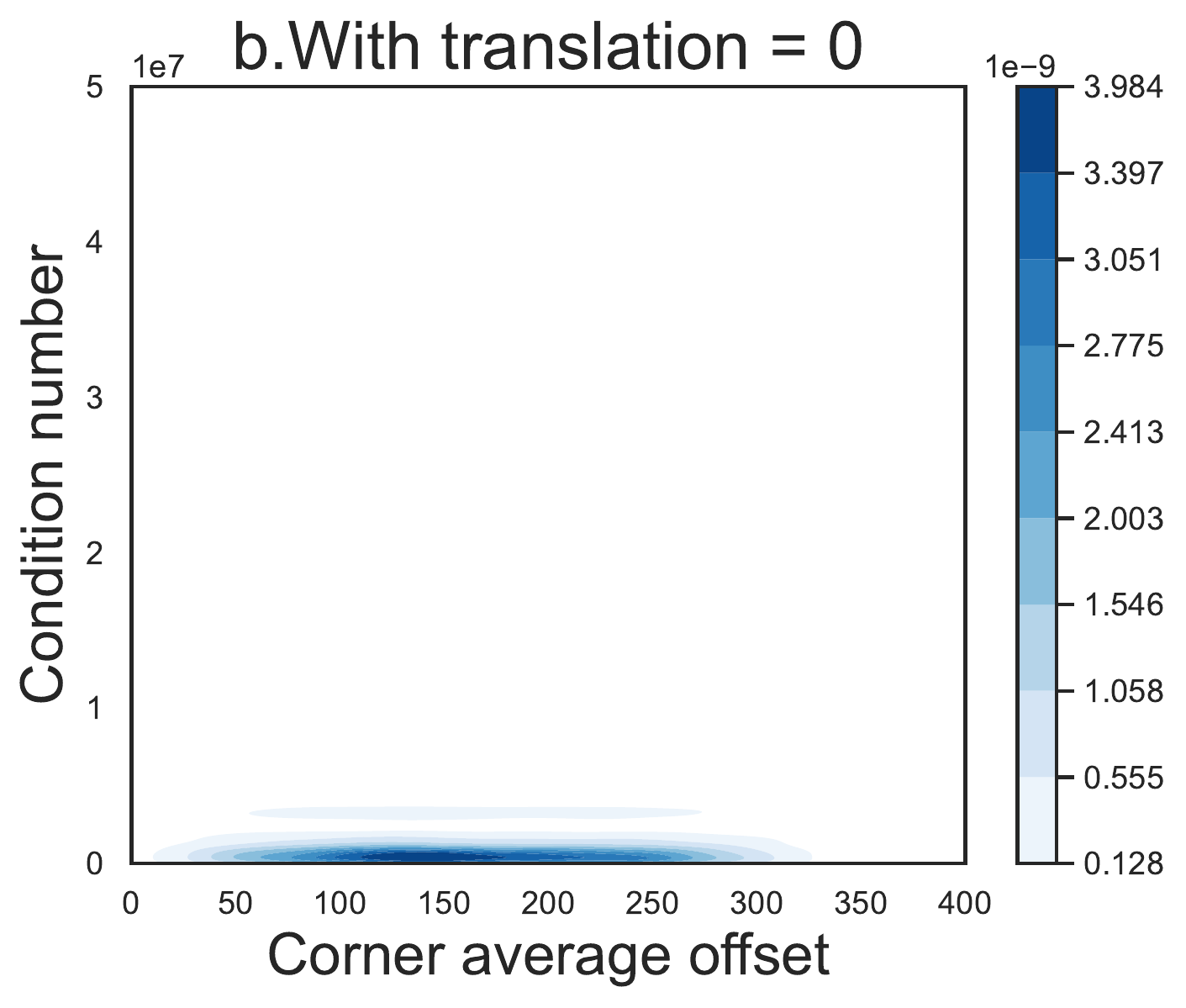}
				%\caption{fig2}
			\end{minipage}
			\label{fig:cond_numb}
		}%
		
		\subfigure{
			\begin{minipage}[t]{0.442\linewidth}
				\centering
				\includegraphics[width=\linewidth]{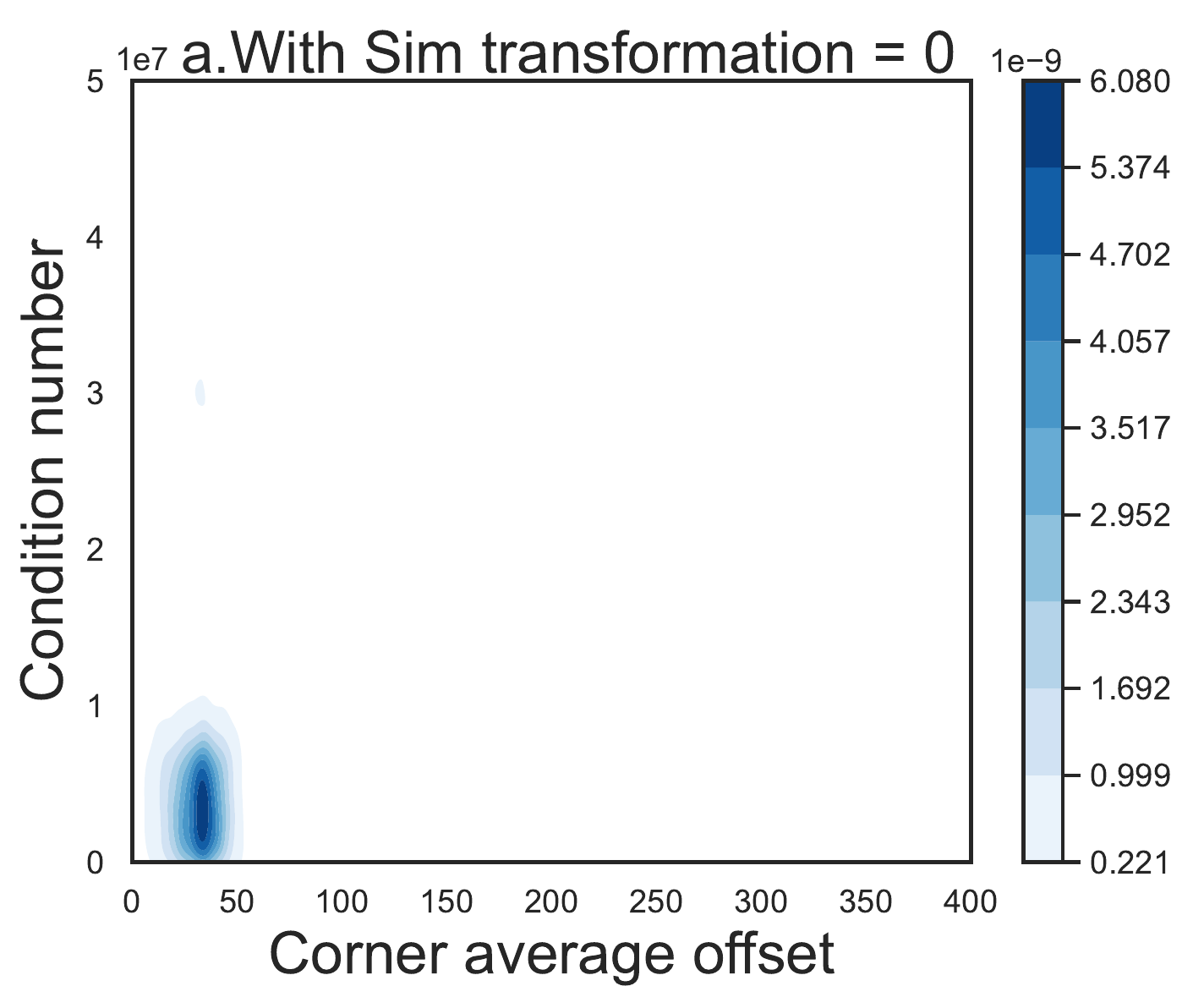}
				%\caption{fig2}
				\label{fig:cond_numc}
			\end{minipage}
		}%trans_vis_cond_d.pdf
		\subfigure{
			\begin{minipage}[t]{0.442\linewidth}
				\centering
				\includegraphics[width=\linewidth]{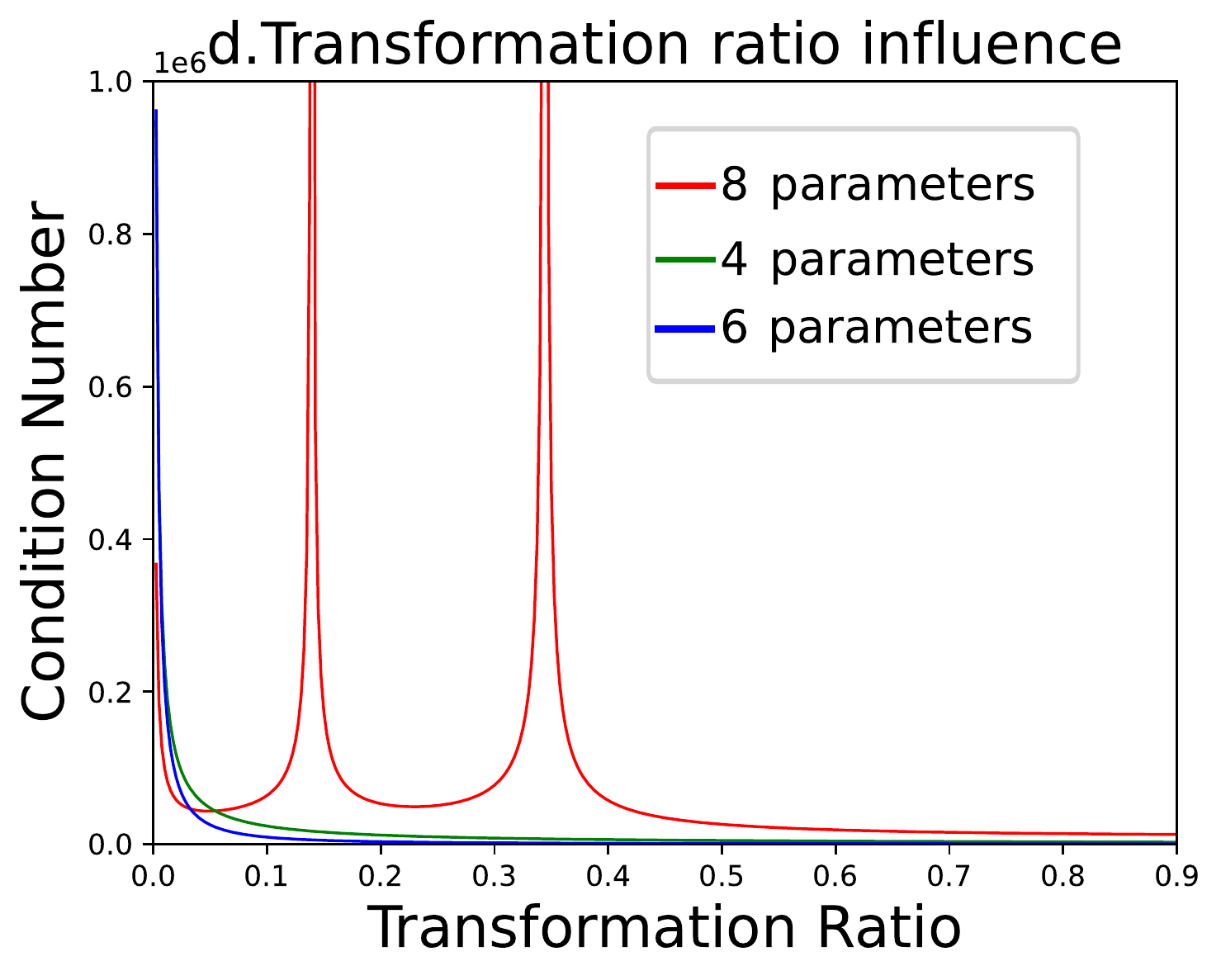}
				%\caption{fig2}
				\label{fig:cond_numd}
			\end{minipage}
		}
		\caption{ (a,b,c) represent the condition number of randomly sampled parameters permute with corner delta, and the value of the right bar represents the probability density of the sampled input point. (d) represents the condition number permute with transformation degree.}
		\label{fig:cond_num}
	\end{figure}

	Fig.~\ref{fig:cond_num} shows the condition number distribution of $\Delta \mathbf{p}(\mathbf{x})$ for a planar transformation.
	We sample $\mathbf{x}$ in the uniform distribution\footnote{All the settings of $\mathbf{x}$ range is detailed in experimental setting} and plot the condition number distribution in different decomposition settings.
	Fig.~\ref{fig:cond_numa} demonstrates that the condition number of directly estimating eight parameters of $\mathbf{H}(\mathbf{x})$ is up to $5e^7$, which makes the system extremely unstable and unreliable for a planar object tracking task.
	We argue that this gives the insight of directly estimating homography parameters is a very hard problem for neural networks.
	On the other hand, Fig.~\ref{fig:cond_numb} and Fig.~\ref{fig:cond_numc} reduce the parameter space by the given translation $[t_1, t_2]$ and similarity $[t_1, t_2, \gamma, \theta]$ parameters, respectively. With these known parameters, the sensitivity drops in a remarkable magnitude, which improves the robustness of the system significantly.
	In addition, Fig.~\ref{fig:cond_numd} shows the change of condition number as all parameters increased by a fixed ratio when large displacement occurs.
	Intuitively, directly estimating eight parameters in a large displacement setting becomes an ill-conditioned problem, however, simply decomposing the transformation contributes more robustness to the system.
	
	Following these observations, we formulate the planar object tracking into a two-step estimation process and decompose the homography into two groups:
	\begin{equation}
	\scriptsize{
		\mathbf{H}(\mathbf{x}) = 
		\begin{bmatrix}
		\gamma\, cos\theta & -\gamma\, sin\theta & t_1 \\
		\gamma\, sin\theta & \gamma\, cos\theta & t_2 \\
		0 & 0 & 1 
		\end{bmatrix}
		\begin{bmatrix}
		k_1 & k_2 & 0 \\
		0 & 1/k_1 & 0\\
		\nu_1 & \nu_2 & 1 
		\end{bmatrix}
		= \mathbf{H}^S \cdot \mathbf{ H}^{\Lambda}
	}
	\label{eq:decompose_matrix}
	\end{equation}
	where $\mathbf{H}^S$ is the similarity transformation, and $\mathbf{H}^\Lambda$ denotes the residual group.
	After decomposition, $\hat{\mathbf{H}}_{k} $ in Eq.~(\ref{eq:total_H}) can be rewritten as $\hat{\mathbf{H}}_{k} = \hat{\mathbf{H}}^S_{k} \cdot
	\hat{\mathbf{H}}^{\Lambda}_{k}$, where $\hat{\mathbf{H}}^S_{k}$ is the estimated similarity transformation from $T$ to $\mathcal{W}(I_k, \mathbf{H}_{k-1}^{-1})$. 
	$\hat{\mathbf{H}}^{\Lambda}_{k}$ is the estimated residual transformation from $T$ to $\mathcal{W}(I_k, (\hat{\mathbf{H}}^S_{k} )^{-1}\cdot \mathbf{H}^{-1}_{k-1})$. In this way, our framework is more stable for estimating by the neural networks.

	\subsection{Homography Decomposition Networks}
	To robustly track the planar object, we introduce novel two-stage end-to-end decomposition networks. Fig.~\ref{fig:all_structure} illustrates the pipeline of our proposed HDN which consists of Similarity Component and Residual Component. They estimate two transformations in order, which are connected by STN~\cite{STN} to make the whole network differentiable. 
	The loss function for our proposed network can be derived as below: 
	\begin{equation}
		\mathcal{L}_{total} = 
		\begin{array}{lr}
			\lambda_1 \mathcal{L}_{s}(\hat{\mathbf{H}}^S) +  \mathcal{L}_{\Lambda}(\hat{\mathbf{H}}
			) 
		\end{array} 
		\label{eq:overall_loss}
	\end{equation}
	where $\mathcal{L}_{s}$ denotes the similarity component loss, $\mathcal{L}_{\Lambda}$ is the residual component loss and $\lambda_1$ is the weight parameter. Our training process only involves two frames due to the formulation of our approach. For simplicity, we remove all the subscript $i$ of $\mathbf{H}_i$ and $I_i$ in the following sections.

	\subsubsection{Similarity Component}
	
\begin{figure}[h]
		\centering 
		\includegraphics[scale=0.5]{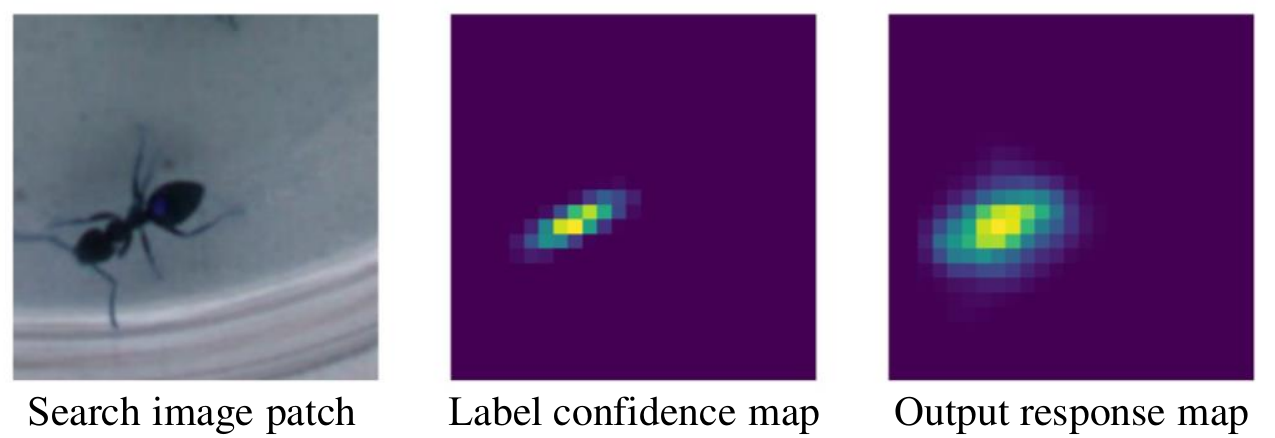}
		\caption{ The classification map of similarity component.}
		\label{fig:label_representation}
	\end{figure}

	As depicted in Fig.~\ref{fig:all_structure}, our similarity component gives reliable initial parameters and does not require to predict a bounding box for the sake of tracking the perspective change. To this end, a classification map and an offset map are generated in the two heads of HDN similar to \cite{SiamBAN} for translation estimation, except that we add more specific object labels adapted to the rotated objects according to the ground-truth $\theta$.
	For the classification map labeling, we simply assign a target possibility for every position in the map and adopts a regression loss to achieve a close probability to the real value. A hamming window is used to construct the label map $\mathbf{M}_c$ as exhibited in Fig.~\ref{fig:label_representation}.
	
	The indices of the highest probability in classification map $\hat{\mathbf{M}}_c$ give the target's center coarsely, since a pixel in the classification map corresponds to eight pixels in the input image due to the total stride $\zeta$ of the network being eight. To achieve an accurate estimation,
	an offset map $\hat{\mathbf{M}}_o$ is thereby utilized to predict the difference between the grid coordinates 
	of the offset map and the real target center location 
	in the search image $I$. 
	Note that, the grid coordinates are aligned to the size of image $I$, and the origin of the coordinates is the center of the object in search image.	
	We choose the max response location $(u_m, v_m) = \mathop{\arg_{u,v}\max} (\hat{\mathbf{M}}_c(u,v))$  in classification map to index the offset map and its grid location $\zeta\cdot(u_m, v_m)$. Therefore, the translation is calculated as $\hat{\mathbf{t}}=(\hat{t}_1, \hat{t}_2) = \hat{\mathbf{M}}_o(u_m,v_m)+\zeta\cdot(u_m, v_m)$.
	
	As in~\cite{cohen2015transformation, Lenc2015UnderstandingIR}, the conventional convolution does not have the equivariance to large scale and rotation changes.  We thereby adopt the warped convolution~\cite{Henriques2017WarpedCE} to extend group convolution to image operator, which obtains an architecture equivariant to arbitrary two-parameter spatial transformations with a proper warping function.
	
	Define a lie group $ \mathcal{G}$ and transformation $ g $ belonging to $ \mathcal{G} $.
	To convert the transformation $g$ into real space, an $\exp$ warp function is adopted to map the element in the lie group~\cite{micro_lie_theory} to scale and rotation operator on a manifold in the real domain. The warped element owns the equivalence to scale and rotation with convolution. The warp function about two parameters scale $\gamma^{\prime}$ and rotation $\theta$ is defined as:
	\begin{equation}
	g_{\gamma^{\prime},\theta}(\mathbf{\tau}) = 
	\begin{bmatrix}
	s^{\gamma^{\prime}} \|\mathbf{\tau}\|\cos(\arctan_2(\tau_2,\tau_1) + \theta) \\
	s^{\gamma^{\prime}} \|\mathbf{\tau}\|\sin(\arctan_2(\tau_2,\tau_1) + \theta)
	\end{bmatrix}
	\label{eq:warp_lp}
	\end{equation}
	where $s$ controls the degree of scaling, $\tau=(\tau_1,\tau_2)$ is the pivot, and $\arctan_2$ denotes standard 4-quadrant inverse tangent
	function.  
	Using Eq.~(\ref{eq:warp_lp}), predicting the scale and rotation could be formulated as a translation estimation problem~\cite{Henriques2017WarpedCE,LDES}.
	
	In this work, Rotation-Scale Equivariant Warping~(RSEW) is proposed to implement the equivariant convolution about scale and rotation. Different from the previous approach PTN~\cite{Esteves2018PolarTN}, our method employs the correlation operator to find the region of interest, where the warped image is further used to estimate the scale and rotation changes by the equivariance with translation.
	%	in the log-polar domain. 
	Let image center be the origin, i.e. the object center of $I$, and left-top point be the resampled image origin. $\gamma'$ is defined as $\frac{2\mu_1}{n}$, and $\theta$ is $\frac{4\pi \mu_2}{n}$. $\mathbf{\mu}=(\mu_1,\mu_2)$ is set as the resampled image coordinates. $s$ in Eq.~(\ref{eq:warp_lp}) is $\frac{n}{4}$, where $n$ is the edge length of squared image $I$. Moreover, we set the pivot $\tau = (1,0)$. The converted grid sampling point $(u, v)$ in $I$ can be represented as follows:
	\begin{equation}
	\left\{
	\begin{aligned}
	&u = \hat{t}_1 + (\frac{n}{4})^{\frac{2\mu_1}{n} } \cos(\frac{4\pi \mu_2}{n}  ) \\
	&v = \hat{t}_2 +  (\frac{n}{4})^{\frac{2\mu_1}{n} } \sin(\frac{4\pi \mu_2}{n}  )
	\end{aligned}
	\right.
	\label{eq:grid_lp}
	\end{equation}
	
	As illustrated in Fig.~\ref{fig:all_structure}, estimating classification and offset maps for the scale and rotation is similar to the translation prediction.
	Once the scale and rotation are obtained, the estimated $\hat{\gamma}$ and $\hat{\theta}$ can be recovered as $(\hat{\gamma}, \hat{\theta})=((\frac{n}{4})^{\frac{2\hat{\mathbf{\mu}}_1}{n}},\frac{4\pi \hat{\mathbf{\mu}}_2}{n})$,
	where $\hat{\mathbf{\mu}}$ is the estimation for scale and rotation in the warped image. 
% 	Its procedure is similar to estimation for $\mathbf{t}$. 
	Composing them and translation estimated before forms similarity 
	parameters $\hat{\mathbf{x}}_S$.
	
	% class-loss-trans：
	Classification loss for translation is based on the weighted label map, which adopts the combined regression loss for the positive samples and negative samples as follows:
	\begin{equation}
		\begin{aligned}
			\mathcal{L}_{cls} = &\frac{1}{K} \sum \limits_{i \in \Omega_{-}}{-\log(1-\hat{\mathbf{M}}_c(i))} \\
			&+\frac{1}{Q}\sum \limits_{j \in \Omega_{+}} {|\hat{\mathbf{M}}_c(j)-\mathbf{M}_c(j)|}\\
		\end{aligned}
		\label{eq:translation_cls_loss}
	\end{equation}	
	where $\Omega_{+}$ denotes the set of top $K$ 
	positive samples in $\hat{\mathbf{M}}_c$. The positive samples are selected when the probability is higher than the given threshold $ \tau$ before choosing the top $K$ samples. 
	To deal with the imbalance between negative and positive samples, $\mathcal{L}_{cls}$ is designed to pick the top $K$ negative samples in order to focus on the hard negatives, and $\Omega_{-}$ represents the chosen negative samples. Q and K are the total quantity of positive and negative samples, respectively.

	For the offset loss $\mathcal{L}_{reg}$, we only calculate the loss of pixels when its positive class probability is higher than the threshold. We use $\mathcal{L}_{reg} = \frac{1}{U} \sum \limits_{i}\mathcal{R}(\hat{\mathbf{M}}_o(i)-{\mathbf{M}}_o(i))$, where $\mathcal{R}$ is the robust loss function (i.e. smooth $l_1$) defined in \cite{FastRCNN}, ${\mathbf{M}}_o$ is the label of offset map, and U denotes the total number of samples in the map.
	The rotation-scale label maps are set similar to  SiamBAN~\cite{SiamBAN}. 
	The classification loss $\mathcal{L}_{cls2}$ for the scale and rotation estimation uses the cross-entropy loss, and the regression employs the same smooth $l_1$ loss as the translation estimation. 
	Thus, the loss of similarity component $L_s$ can be derived as below:
	\begin{equation}
	\mathcal{L}_{s} = \mathcal{L}_{cls} + \mathcal{L}_{reg}+ \mathcal{L}_{cls2} + \mathcal{L}_{reg2}
	\label{eq:sim_total_loss}
	\end{equation}
	
	\subsubsection{Residual Component}
	
	Residual transformation estimation is subsequently accomplished by 
	regressing the corners offsets $\Delta \mathbf{P}$ between the tracked target quadrilateral $(\hat{\mathbf{H}}^{S})^{-1}\mathbf{P_1}$ and  $\mathbf{P}_T $ as introduced in Fig.~\ref{fig:all_structure}, where $\mathbf{P}_1$ is the object corners coordinates in the search patch. The prediction $\Delta \hat{\mathbf{P}}$ is converted to the homography $\mathbf{\hat{H}}^{\Lambda}$ by DLT~\cite{DLT}. Inspired by~\cite{Nguyen2018UnsupervisedDH, Zhang2020ContentAwareUD}, a fused semi-supervised network is adopted to estimate the homography between two images, which makes the whole network differentiable through STN~\cite{STN}. The reason of using the semi-supervised setting is that many tracking datasets are not labeled with homography and the simple augmentation for the supervised setting cannot cover the appearance changes in real-world scenarios. 
	The loss for residual homography estimation employs a triplet loss $\mathcal{L}_{\Lambda+}^*=\mathcal{L}_{Triplet}$ defined in \cite{FaceNet}, where the embedding matrices are the anchor embedding feature $E(I)$, the positive embedding $E(\mathcal{W}(T, \hat{\mathbf{H}}^S \cdot \hat{\mathbf{H}}^\Lambda))$, and the negative embedding $E(T)$. $E$ is the coarse feature extractor as shown in Fig.~\ref{fig:all_structure}.
	Concretely, $\mathcal{L}_{\Lambda+}^*$ can be written as follows:
% 	\begin{equation}
% 	    \begin{aligned}
%     		\mathcal{L}_{\Lambda+}^* =\frac{1}{n^2}\sum_j {max&\{\Vert E(I_1)_j - E(\mathcal{W}(I_2, (\hat{\mathbf{H}}^S_{i})^{-1}))_j\Vert_2\ \\
%     		&- \Vert E(I_1)_j- E(I_2)_j \Vert_2+\alpha, 0\}} \\
% 		\end{aligned}
% 		\label{eq:unsup_res_loss}
% 	\end{equation}	
	\begin{equation}
    \begin{aligned}
	\mathcal{L}_{\Lambda+}^* =\frac{1}{m^2}\max &\{\Vert E(I) -E(\mathcal{W}(T,\hat{\mathbf{H}}^S \cdot \hat{\mathbf{H}}^{\Lambda}))\Vert_2\ \\ & - \Vert E(T)- E(I)\Vert_2+\alpha,\mathbf{0}\}
    \end{aligned}
	\label{eq:unsup_res_loss}
	\end{equation}
	where $m$ denotes the edge length of the embedding, and $\alpha$ is set to one by default. Pairwise distance with $l_2$-norm is adopted in calculating the distance of maps. 
	
	To adapt the synthetic training data and acquire higher estimation accuracy, we add an additional supervised loss with $l_1$-norm for the object corners' offsets as  
    $
	\mathcal{L}_{\Lambda+} = \frac{1}{4} \sum_{i}^4 ||\Delta \mathbf{P}(i) - \Delta \hat{\mathbf{P}}(i)||_1$.

	 The negative samples of two different objects are added to further improve the robustness of our model, and we employ the strategy to set corners' translation to zero. 
    We define the homography estimation of a negative sample loss as
	$\mathcal{L}_{\Lambda-} =  \frac{1}{4}\sum_{i}^4 \mathcal{R}(\Delta \hat{\mathbf{P}}(i)).$
	Intuitively, this strategy guides the tracker staying on its previous status when it loses the target.
	 
	From the above all, the total loss $\mathcal{L}_{\Lambda}$ of residual component can be obtained as below:
	\begin{equation}
	\mathcal{L}_{\Lambda} = \lambda_2 \mathcal{L}_{\Lambda-} + \lambda_3 \mathcal{L}_{\Lambda+}^{*}+\lambda_4 \mathcal{L}_{\Lambda+}
	\label{eq:hm_total_loss}
	\end{equation}
	where $\lambda_2$, $\lambda_3$ and $\lambda_4$ are the weights to balance the loss. 
	Our proposed method with robust similarity prediction and the constrained negative loss bridges the gap between homography estimation and planar tracking. 
	
	\begin{figure*}[htbp]
		\centering
		\subfigure{
			\begin{minipage}[t]{0.26\linewidth}
				\centering
				\includegraphics[width=1\linewidth]{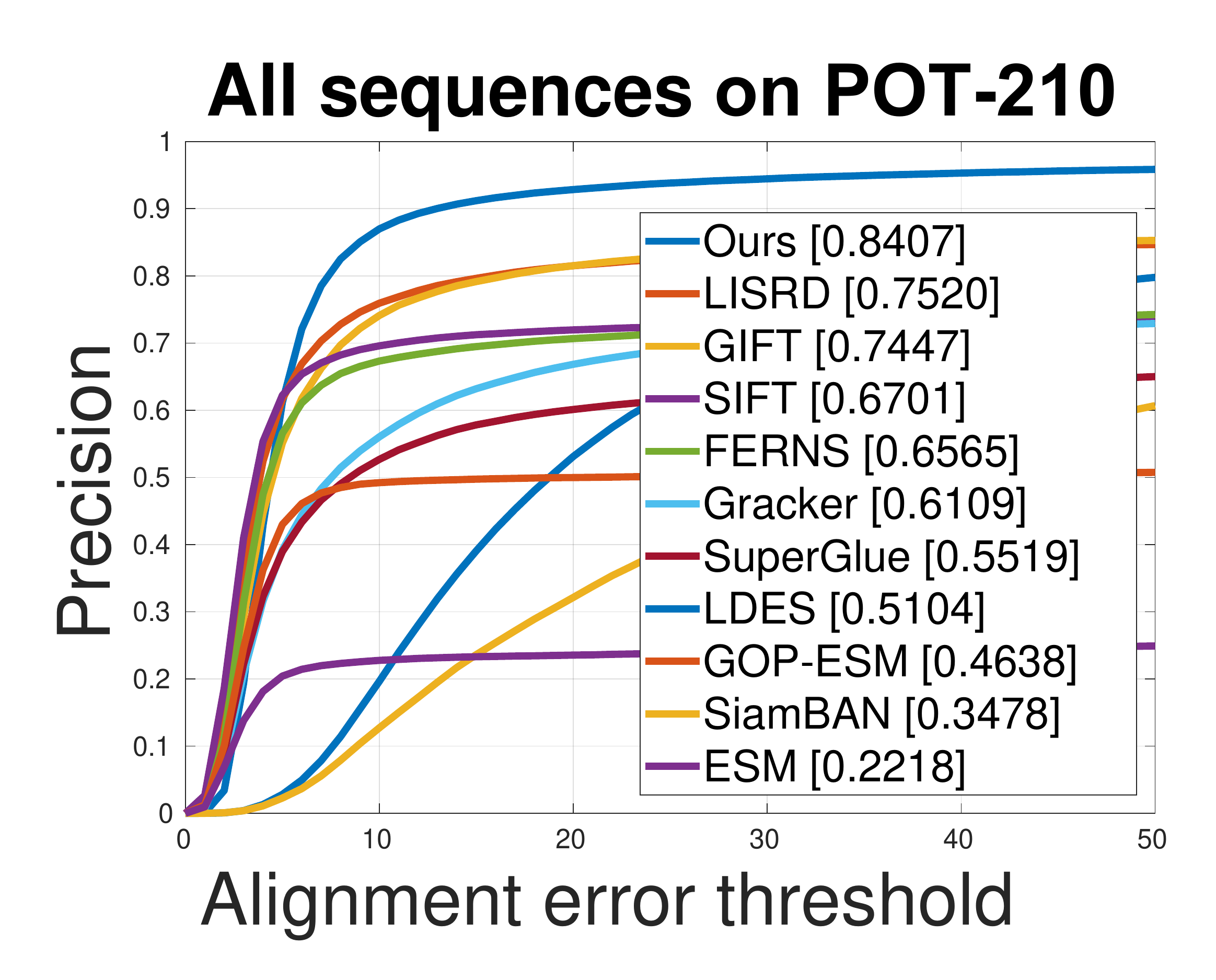}
				%\caption{fig1}
			\end{minipage}%
		}%
		\subfigure{
		\begin{minipage}[t]{0.26\linewidth}
			\centering
			\includegraphics[width=1\linewidth]{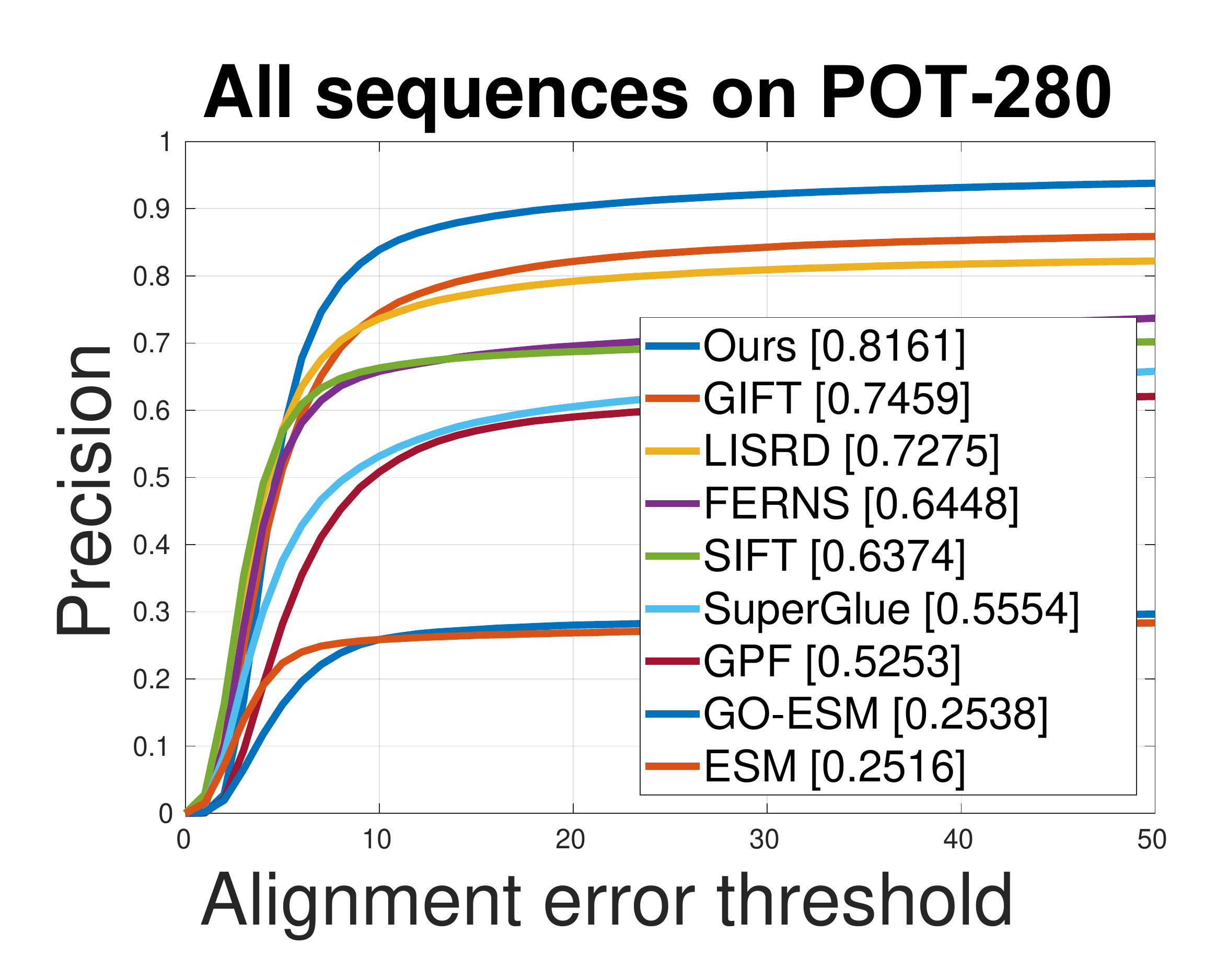}
			%				\caption{fig2}
		\end{minipage}
		}%
		\centering
		\subfigure{
			\begin{minipage}[t]{0.29\linewidth}
				\centering
				\includegraphics[width=1\linewidth]{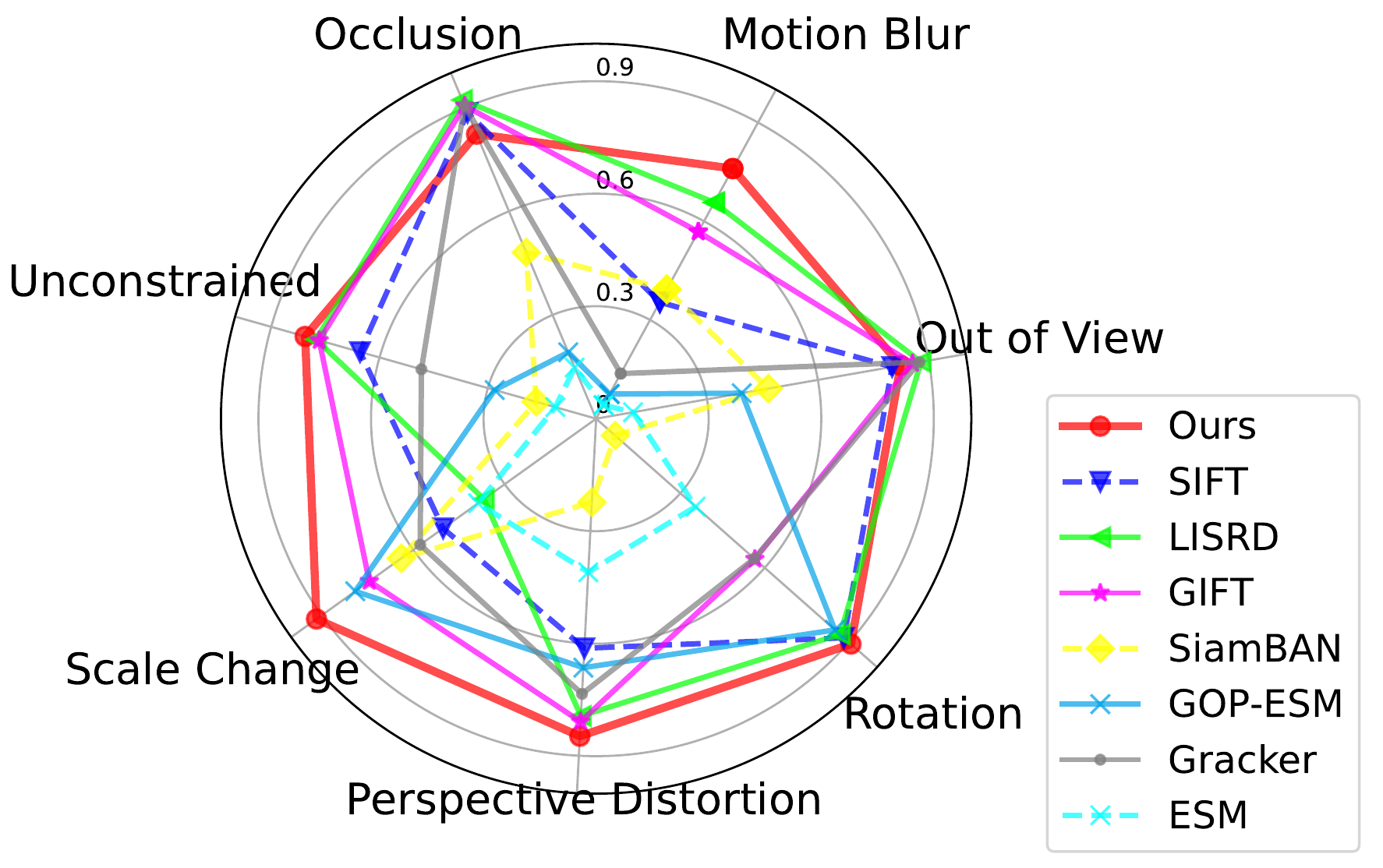}
				%\caption{fig2}
			\end{minipage}
		}%
		
		\centering
		\caption{Comparisons on POT-210 and POT-280 with various challenging factors.
			The left subfigure and middle subfigure are the Precision on POT-210~\cite{POT} and POT-280~\cite{POT280} with different thresholds, respectively. Legends show the average precision (avg Prec). The third radar figure compares the trackers' avg Prec on 7 challenging factors}
		\label{fig:pot_compare}
%		\vspace{-0.2in}
	\end{figure*}
	\subsection{Training Details}
	Existing datasets lack the transformation parameters including rotation and scale, etc. Therefore, we augment the possible transformations according to Eq.~(\ref{eq:decompose_matrix}). The algorithm randomly chooses $\mathbf{x}$ from a certain range to perform the transformation on COCO14~\cite{COCO}. The range is smaller than the whole domain because our proposed compositional method accumulates the transformation between inter-frames. To solve the unrealistic problem of synthetic datasets for residual component training, we sample the images of tracking dataset GOT10k~\cite{GOT10k} with a small interval threshold as training data and adopt an unsupervised residual loss $\mathcal{L}_{\Lambda+}^*$. We further discuss the effect of the supervision method on the ablation section.
	
% 	\textcolor{blue}{We use ResNet50 and ResNet34 as the backbones respectively in stage one and stage two.}
	
% 	The backbone of the first stage network is a reduced stride ResNet50 pre-trained on ImageNet, where we choose C3, C4, and C5 layers features as our input same as~\cite{SiamBAN}, the stride in conv4 and conv5 blocks are set to one, the atrous rate is two and four, respectively. The second stage backbone is a regular ResNet34 pre-trained model, and we change its input into two channels. ResNet50 is unchanged during training while the ResNet34 is trained from scratch with a small learning rate.
	
	\section{Experiment}
	\subsection{Experiment Setup}
	We conducted experiments on a PC with an intel E5-2678-v3 processor~(2.5GHz), 32GB RAM and Nvidia GTX 2080Ti GPU. Our proposed method is implemented in Pytorch. The size of input template $T$ for our networks is $127\times 127$, while search image $I$ has the size of $255\times 255$ to deal with the large homography changes. All the hyperparameters are set empirically, and we do not use any re-initialization and failure detection scheme. For the hyperparameters of HDN in training and testing, we set 	
	$\lambda_1=100$ , $\lambda_2=1$, $\lambda_3=1$, $\lambda_4=0.25$, $K=100$.
	$\gamma \in [1/1.38, 1.38]$, $\theta \in [-0.7, 0.7]$, $t\in[-32,32]$, $k_1\in[-0.1, 0.1]$, $k_2 \in [-0.015,0.015]$, $\nu \in [-0.0015,0.0015]$. The probability threshold $\tau$ in $\mathcal{L}_{cls}$ is set to 0.7.

	\subsection{State-of-the-art Comparison}
	We compare our proposed method with the state-of-the-art trackers on four tracking datasets. These methods can be divided into three categories, including keypoint-based trackers (SIFT~\cite{SIFT}, SURF~\cite{SURF}, FERNS~\cite{FERNS}, Gracker~\cite{Gracker}, SuperGlue~\cite{SuperGlue}, LISRD~\cite{LISRD}, GIFT~\cite{GIFT}), region-based method (ESM~\cite{ESM}, GOP-ESM~\cite{GOP-ESM}) and generic visual tracking method (GPF~\cite{GPF}, Ocean~\cite{Ocean}, TransT~\cite{TransT} and SiamBAN~\cite{SiamBAN}). 
	We evaluate all the trackers on the challenging POT-210 dataset and choose top-ranked trackers in three categories for the POT-280, UCSB, and POIC. All the results are either based on the publications or from the benchmark~\cite{POT, POT280}. 
	
	\begin{table}[tbp]
		\centering
		\resizebox{\linewidth}{!}{
			\begin{tabular}{p{4.59em}|p{5.08em}ccccc}
				\hline
				\textbf{tracker} & \textbf{\makecell[c]{avg\\ Prec/HSR}} & \multicolumn{1}{p{3.72em}}{\textbf{\makecell[c]{Prec\\ (e$\leq$ 5)}}} & \multicolumn{1}{p{3.835em}}{\textbf{\makecell[c]{Prec\\ (e$\leq$ 10)}}} & \multicolumn{1}{p{3.39em}}{\textbf{\makecell[c]{Prec\\ (e$\leq$ 20)}}} & \multicolumn{1}{p{3.39em}}{\textbf{\makecell[c]{avg\\ CP}}} & \multicolumn{1}{p{3.665em}}{\textbf{\makecell[c]{avg\\ SR}}} \\
				\hline
				SIFT  & 0.670/0.570 & \textcolor[rgb]{ 1,  0,  0}{\textbf{0.622 }} & 0.695  & \textcolor[rgb]{ .329,  .51,  .208}{\textbf{0.719 }} & 0.698  & 0.701  \\
				SURF  & 0.657/0.536 & 0.543  & 0.663  & 0.711  & 0.689  & 0.705  \\
				LISRD & \textcolor[rgb]{ .184,  .459,  .71}{\textbf{0.752/0.619}} & \textcolor[rgb]{ .184,  .459,  .71}{\textbf{0.617 }} & \textcolor[rgb]{ .184,  .459,  .71}{\textbf{0.759 }} & \textcolor[rgb]{ .184,  .459,  .71}{\textbf{0.815 }} & \textcolor[rgb]{ .184,  .459,  .71}{\textbf{0.788 }} & \textcolor[rgb]{ .184,  .459,  .71}{\textbf{0.805 }} \\
				GIFT  & \textcolor[rgb]{ .329,  .51,  .208}{\textbf{0.745}}/0.580 & 0.551  & \textcolor[rgb]{ .329,  .51,  .208}{\textbf{0.741 }} & \textcolor[rgb]{ .184,  .459,  .71}{\textbf{0.815 }} & \textcolor[rgb]{ .329,  .51,  .208}{\textbf{0.785 }} & \textcolor[rgb]{ .329,  .51,  .208}{\textbf{0.801 }} \\
				SuperGlue & 0.552/0.509 & 0.389  & 0.526  & 0.601  & 0.592  & 0.659  \\
				Gracker & 0.611/0.527 & 0.392  & 0.560  & 0.668  & 0.684  & 0.716  \\
				ESM   & 0.222/0.209 & 0.204  & 0.227  & 0.235  & 0.242  & 0.261  \\
				FERNS & 0.657/\textcolor[rgb]{ .329,  .51,  .208}{\textbf{0.601}} & 0.565 & 0.673  & 0.706  & 0.686  & 0.724  \\
				SCV   & 0.250/0.233 & 0.228  & 0.257  & 0.265  & 0.274  & 0.287  \\
				GOP-ESM & 0.464/0.442 & 0.430  & 0.492  & 0.499  & 0.484  & 0.497  \\
				SiamBAN & 0.348/0.335 & 0.022  & 0.127  & 0.321  & 0.650  & 0.693  \\
				LDES  & 0.510/0.518 & 0.028  & 0.196  & 0.531  & 0.617  & 0.737  \\
				Ocean & 0.255/0.261 & 0.011  & 0.089  & 0.241  & 0.464  & 0.521  \\
				TransT & 0.464/0.381 & 0.090  & 0.276  & 0.486  & 0.764  & 0.774  \\
				\hline
				\textbf{Ours} & \textcolor[rgb]{ 1,  0,  0}{\textbf{0.841/0.710}} & \textcolor[rgb]{ .329,  .51,  .208}{\textbf{0.611 }}  & \textcolor[rgb]{ 1,  0,  0}{\textbf{0.870}} & \textcolor[rgb]{ 1,  0,  0}{\textbf{0.928 }} & \textcolor[rgb]{ 1,  0,  0}{\textbf{0.886 }} & \textcolor[rgb]{ 1,  0,  0}{\textbf{0.904 }} \\
				\hline
			\end{tabular}%
		}
		\caption{Results on POT-210. Top-3 results of each dimension are colored in red, blue and green, respectively.}
		\label{tab:POT_results}
	\end{table}%
	
	\textbf{POT}. 
	POT-210~\cite{POT} contains 210 videos of 30 planar objects sampled in the natural environment. It includes seven challenging scenarios: scale, rotation, occlusion, etc. POT-280~\cite{POT280} adds another 70 sequences to POT-210, . 
	and adopt two evaluation metrics: Precision~(Prec) and Homography Success Rate~(HSR). Precision is defined as the percentage of frames whose alignment error is smaller than the given threshold. The alignment error is computed by the average of the four points $L_2$ distance between the predicted polygon and the ground truth label. The Homography Success Rate~(HSR) describes the percentage of frames whose homography discrepancy score is less than a threshold.  
	Fig.~\ref{fig:pot_compare} reports the result, which indicates that our HDN outperforms the other trackers in most of the metrics on both benchmarks.
	We give more experiment results in the appendix. Using the relative improvement ratio, it achieves $12\%$ and $9.4\%$ improvement compared to the second-best methods in avg Prec on POT-210 and POT-280,
	respectively. Moreover, our presented HDN method performs better than the best non-keypoint-based tracker
	LDES~\cite{LDES} by $64.7\%$ on POT-210. 
	Our decomposition deep networks can robustly estimate the homography in all challenging scenarios. For some hard cases~(see the right subfigure in Fig.~\ref{fig:pot_compare}), our method performs better than other trackers except for the occlusion and out-of-view scenes due to the correlation mechanism. 
	The reason is that the center of its response map is estimated from the non-occluded region if an object is heavily occluded.
	LISRD~\cite{LISRD} and GIFT~\cite{GIFT} have difficulties in dealing with scale variations or rotation changes. 
	Shrinking object reduces the total number of keypoints, which makes the result of homography estimation unreliable. 
	In terms of perspective transformation, our method performs better than the keypoint-based methods. This is mainly because they estimate from the template to the current frame to avoid failure while incurring the jitters at the same time. 
	
	To evaluate the quality of trackers from other aspects and reveal the underlying reasons, we include two extra metrics. 
	Centroid Precision~(CP) is the precision of the object's polygon centroid, and Success Rate~(SR) is defined as the successfully tracked ratio with the overlap (Intersection over Union) greater than the given ratio. 
% 	The average Prec, average CP, average HSR, and average SR denote the average~(avg) value traversing each given possible threshold. 
	Table~\ref{tab:POT_results} lists more trackers and detailed results on POT-210. HDN exhibits a great improvement on almost all the metrics, where its average Prec and HSR are much higher than the second-best method. Average Centroid Precision manifests our higher accuracy of translation estimation and average SR exhibits the better overlap score. When the error threshold is small, our method is not as good as other conventional trackers. Unlike other trackers, we only apply HDN once, ESM-based method iteratively optimizes the estimation with more accurate results in easy cases. 
    The input image size of HDN is smaller than most of the conventional methods. In the extreme case, there is an error of 10 pixels in resolution. %Practically, errors under 5 pixels can be neglected in real-world applications. 
    Besides, unsupervised learning may lead to inferior precision results in the case of low error thresholds. This is mainly because the loss of unsupervised learning may not be directly related to the corner errors.

	\begin{table}[htbp]
		\centering
		\resizebox{\linewidth}{!}{
			\begin{tabular}{c|c|ccccc}
				\hline
				\multicolumn{2}{c|}{\textbf{Trackers}} & \textbf{SIFT} & \textbf{Gracker} & \textbf{TransT} & \textbf{GOP-ESM} & \textbf{Ours} \\
				\hline
				POIC  & avg Prec & 0.527 & 0.819 & 0.367 & \textcolor[rgb]{ 0,  .439,  .753}{\textbf{0.873}} & \textcolor[rgb]{ 1,  0,  0}{\textbf{0.874}} \\
				& Prec(e$\leq$ 5) & 0.400  & 0.671 & 0.047 & \textcolor[rgb]{ 1,  0,  0}{\textbf{0.868}} & \textcolor[rgb]{ 0,  .439,  .753}{\textbf{0.749}} \\
				& Prec(e$\leq$ 10) & 0.520 & 0.829 & 0.185 & \textcolor[rgb]{ 0,  .439,  .753}{\textbf{0.893}} & \textcolor[rgb]{ 1,  0,  0}{\textbf{0.894}} \\
				& Prec(e$\leq$ 20) & 0.573 & 0.878 & 0.366 & \textcolor[rgb]{ 0,  .439,  .753}{\textbf{0.901}} & \textcolor[rgb]{ 1,  0,  0}{\textbf{0.948}} \\
				& avg CP  & 0.547 & 0.870  & 0.722 & \textcolor[rgb]{ 0,  .439,  .753}{\textbf{0.889}} & \textcolor[rgb]{ 1,  0,  0}{\textbf{0.916}} \\
				& avg SR & 0.570  & 0.871 & 0.709 & \textcolor[rgb]{ 0,  .439,  .753}{\textbf{0.882}} & \textcolor[rgb]{ 1,  0,  0}{\textbf{0.923}} \\
				\hline
				UCSB  & avg Prec & \textbackslash{} & \textcolor[rgb]{ 0,  .439,  .753}{\textbf{0.837}} & 0.422 & 0.528 & \textcolor[rgb]{ 1,  0,  0}{\textbf{0.871}} \\
				& Prec(e$\leq$ 5) & \textbackslash{} & \textcolor[rgb]{ 0,  .439,  .753}{\textbf{0.648}} & 0.000  & 0.487 & \textcolor[rgb]{ 1,  0,  0}{\textbf{0.660}} \\
				& Prec(e$\leq$ 10) & \textbackslash{} & \textcolor[rgb]{ 0,  .439,  .753}{\textbf{0.831}} & 0.001 & 0.519 & \textcolor[rgb]{ 1,  0,  0}{\textbf{0.916}} \\
				& Prec(e$\leq$ 20) & \textbackslash{} & \textcolor[rgb]{ 0,  .439,  .753}{\textbf{0.903}} & 0.141 & 0.552 & \textcolor[rgb]{ 1,  0,  0}{\textbf{0.964}} \\
				& avg CP  & \textbackslash{} & 0.885 & \textcolor[rgb]{ 0,  .439,  .753}{\textbf{0.908}} & 0.575 & \textcolor[rgb]{ 1,  0,  0}{\textbf{0.912}} \\
				& avg SR & \textbackslash{} & \textcolor[rgb]{ 0,  .439,  .753}{\textbf{0.859}} & 0.565 & 0.579 & \textcolor[rgb]{ 1,  0,  0}{\textbf{0.887}} \\
				\hline
			\end{tabular}%
		}
		\caption{Results on UCSB and POIC. Top-2 results of each dimension (row) are colored in red and blue, respectively.}
		\label{tab:others-dat-compare}%
	\end{table}%
	
	\textbf{UCSB, POIC}.
	We further conduct experiments on another two challenging datasets UCSB~\cite{UCSB} and POIC~\cite{GOP-ESM}.
% 	UCSB~\cite{UCSB} consists of 96 videos with six different texture planar targets having a total of 6,889 frames, including transformation such as panning, zooming, tilting, and rotation. It has 9 levels of motion blurs with different illumination conditions. POIC~\cite{GOP-ESM} is a planar tracking dataset for objects with illumination changes that contain 20 video sequences~(22,971 frames). 
	Table~\ref{tab:others-dat-compare} 
	reports the results compared to other trackers. Note that the keypoint-based methods such as LISRD and GIFT are not designed for planar tracking. There are no published results on UCSB and POIC.  Our approach performs the best on all the metrics on both datasets, except when the error is smaller than 5 pixels in POIC. The reason behind this is the same as on POT. The performance of SIFT is inferior to the region-based methods, which may due to the low texture quality, motion blurs, and illumination changes.

	% 	\subsection{AR application}
	% 	To show our robustness importance, we implement a naive demo to replace the advertisement board with an advertisement video. And compare different trackers performance as shown in Fig.~\ref{fig:replacement_comparision}, our method gives a neat and strong estimation, while others such as SIFT~\cite{SIFT} failed in a few frames. Our baseline SiamBAN~\cite{SiamBAN} generates the video fakely although with robustness. \textcolor{red}{Have not found the appropriate video as background. So currently I used POT-210 as the background}

\begin{figure}[t]
		\subfigure{
			\begin{minipage}[t]{0.47\linewidth}
				\centering
				\includegraphics[width=1\linewidth]{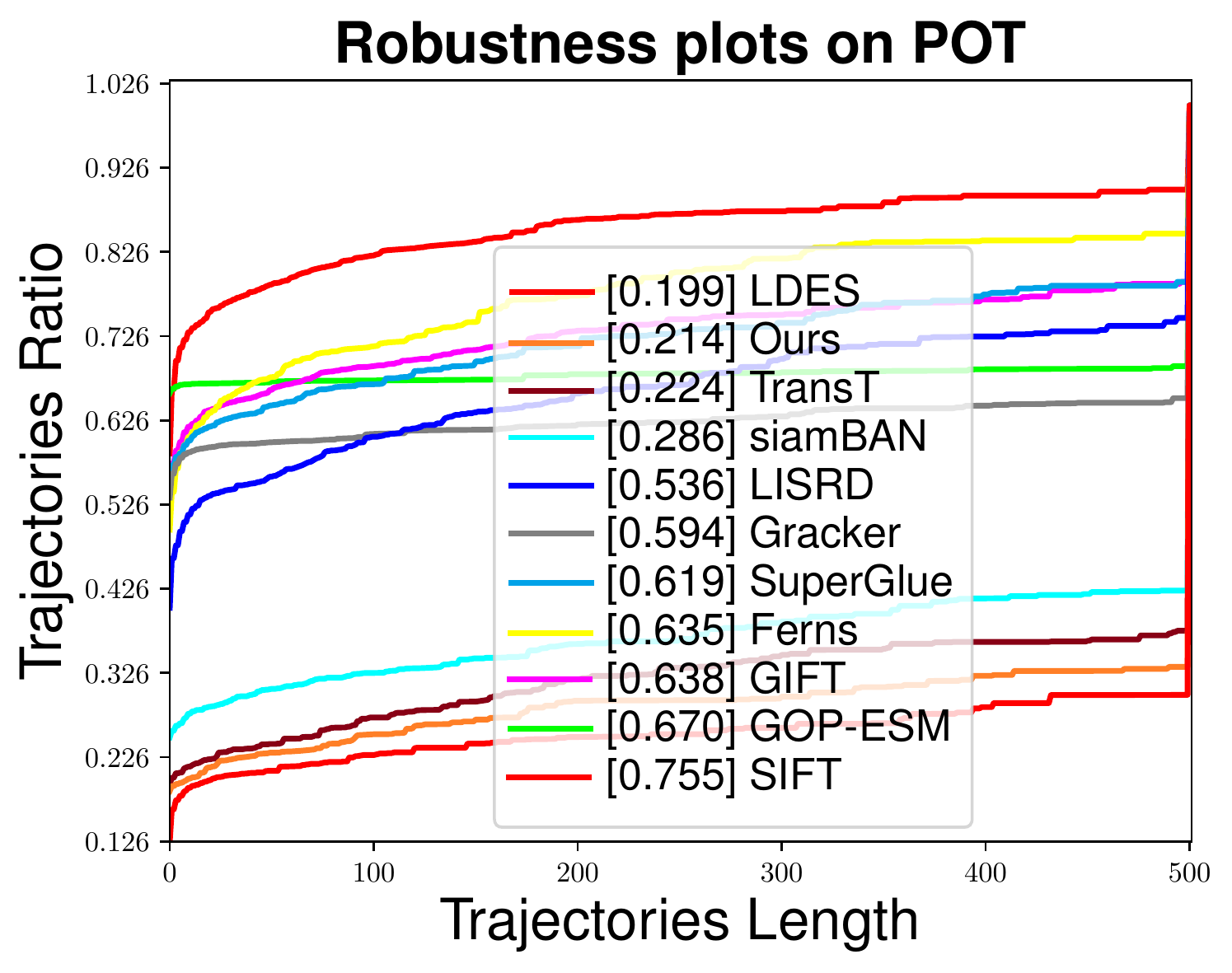}
				%\caption{fig1}
			\end{minipage}%
		}%
		\subfigure{
			\begin{minipage}[t]{0.49\linewidth}
				\centering
				\includegraphics[width=1\linewidth]{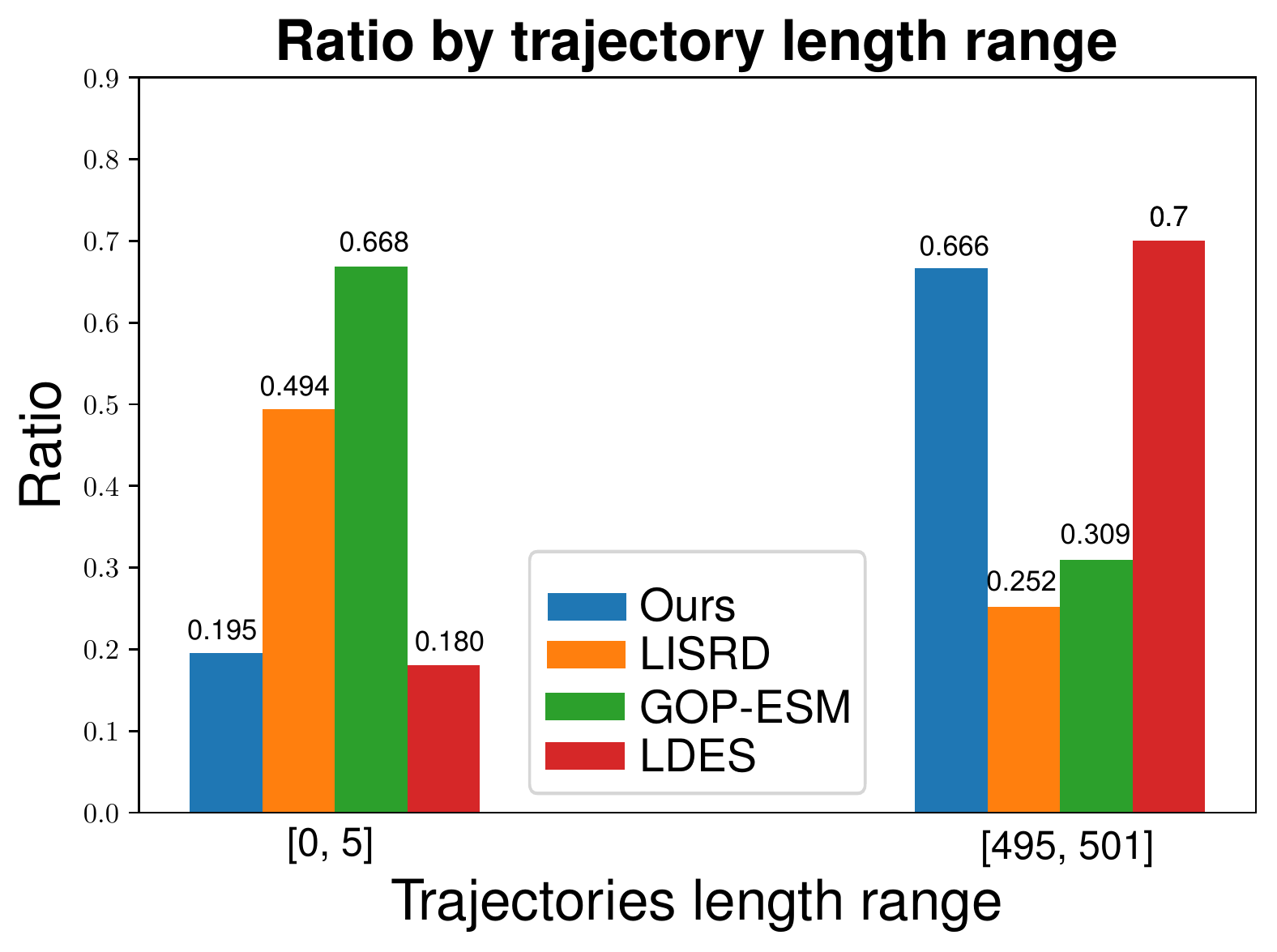}
				%\caption{fig2}
			\end{minipage}
		}%
		\caption{Tracking robustness evaluation on POT. The right subfigure is trajectories range ratio with length range [0,5] and [495, 501]. }
		\label{fig:POT_robustness}
% 		\vspace{-0.2in}
	\end{figure}

	\paragraph{Tracking Robustness}
	To further evaluate the robustness of the tracker, we plot the trajectory robustness which is the ratio of trajectories with different lengths. It is defined as the length having no failure within the threshold IoU $>$ 0.2. We set the trajectory length threshold to 10 for comparison in the legend of Fig.~\ref{fig:POT_robustness}. All the keypoint-based methods generate large trajectory fragments, which means they are easy to lose their target. Although SIFT achieves a high average Prec in POT, it has the lowest robustness with the highest fragments ratio $75.5\%$ in the case of trajectories length less than 10. This cannot be ignored because the frequent lost of objects brings the unsatisfied experiences, especially in augmented reality applications. HDN and LDES have fewer short trajectories and more trajectories cover the whole sequence than LISRD and GOP-ESM. It reveals that predicting translation first is crucial for improving the robustness of planar tracking.

    \subsection{Ablation Study}
	
	We study the impact of individual components in HDN, and conduct the ablation on POT-210, as reported in Table~\ref{tab:ablation}.
	To evaluate the contribution of different stages, we separate stages in HDN, and No.(1,2,8) give their performance. With only similarity component (Sim), the average Prec of HDN is higher than SIFT. On the other hand, only using the residual component (Res) leads to a rapid drop of the average Prec, which is only $31.9\%$. It coincides with the observation in Fig.~\ref{fig:cond_num} and shows the effectiveness of our proposed homography decomposition networks. HDN is efficient in practice, which runs at 10.6 fps.
	To be unsupervised or not, that is a question in learning. However, we find they are not conflicted. No.(3,4,8) show that using both the supervised and unsupervised loss in residual component obtains the higher average Prec than the single loss. %It may be from our unchanged structure, we only add more data to let the Residual component learn more.
	
	\begin{table}[t]
		\centering
		\resizebox{\linewidth}{!}{
			\begin{tabular}{|c|l|l|ccc|cc|}
				\hline
				\multicolumn{1}{|l|}{\textbf{No.}} & \textbf{Component} & \textbf{Sup-method} & \multicolumn{1}{l}{ $ \mathcal{L}_{cls} \& \mathcal{L}_{loc}$   } & \multicolumn{1}{l}{\textbf{Rot-label}} & \multicolumn{1}{l|}{\textbf{Neg-loss}} & \textbf{Prec} & \textbf{Speed (fps)} \\
				\hline
				1     & Sim   & both  &   \cmark    &  \cmark     & \cmark  &   0.689    & 13.7 \\
				2     & Res   & both  & \cmark      &  \cmark     & \cmark  &    0.319   & 21.2 \\
				\hline
				3     & Sim+Res & Supervised &  \cmark   & \cmark     & \cmark      & 0.837     & \textbackslash{}  \\
				4     & Sim+Res & Unsupervised & \cmark       & \cmark      &   \cmark    & 0.417    & \textbackslash{} \\
				\hline
				5     & Sim+Res & both  &      & \cmark     &  \cmark &  0.732   & \textbackslash{} \\
				6     & Sim+Res & both  &   \cmark    &     &  \cmark     & 0.836  & \textbackslash{} \\
				7     & Sim+Res & both  & \cmark      &  \cmark     &      & 0.733  & \textbackslash{} \\
				\hline
				8     & Sim+Res & both  &   \cmark    & \cmark      &  \cmark     & 0.841  & 10.6 \\
				\hline
			\end{tabular}%
		}
		\caption{Ablation on different components, Residual component supervision method, other designs and speed testing. }
		\label{tab:ablation}%
	\end{table}%

	%	\textbf{Unsupervised}

	We further discuss the effect of other aspects. A rotated classification map label (No.6) brings a little improvement. This is because the rotated label is useful for the occluded object, where the offset estimation is inaccurate. Our proposed $\mathcal{L}_{cls}$ and $\mathcal{L}_{reg}$ is crucial in our HDN as a result of the relief of positive-negative samples imbalance and hard negative sampling problem. We combine them in Table~\ref{tab:ablation}, as they are related. The average Prec descends to $73.2\%$ without them. Negative loss is crucial in the residual component, and the average Prec drops to $73.2\%$ without it. Large appearance changes typically occur during tracking, and a compositional method may fail due to the large estimation error in the single frame. 

	\section{Conclusion}
	In this paper, we have proposed novel homography decomposition networks that drastically reduce and stabilize the condition number by decomposing the homography transformation into two groups. Specifically, a similarity transformation estimator was designed to predict the first group robustly by a deep convolution equivariant network. By taking advantage of the scale and rotation estimation with high confidence, a residual transformation was estimated by a simple regression model. Furthermore, we presented an end-to-end semi-supervised training scheme to effectively capture the homography transformations. Extensive experiments show that our proposed approach outperforms the state-of-the-art planar tracking methods at a large margin on the challenging POT, UCSB and POIC datasets.
	
	\section{Acknowledgement}
	This work is supported by the National Natural Science Foundation of China under Grants (61831015 and 62102152) and sponsored by CAAI-Huawei MindSpore Open Fund.

	\bibliography{HDN}

	\clearpage
    \appendix
    {\noindent\Large\textbf{Appendix}}
    \newline
    
	In this appendix, we give more detailed information on our proposed Homography Decomposition Netowrks approach. Firstly, we describe the definition of condition number in our presented method. Secondly, we give more details on our experimental implementation. We then include the additional visual comparisons and the quantitative results in the different scenarios, and discuss the failure cases of our proposed method. 
	
	\section{Condition Number}
	Condition number of a function indicates how much the output value of a function can be affected by the changes of the input argument. Directly regressing four corner points is a possible deep homography estimation method for planar object tracking. Points movements do not reflect the real object transformation, we thereby use another parameterization method to represent the points offsets as introduced in Section "Decomposition". Corner offsets regarding the parameters $\mathbf{x}$ are $
	\Delta \mathbf{p}(\mathbf{x}) = \mathbf{H}(\mathbf{x})\cdot \mathbf{p}-\mathbf{p}$.
	The condition number $Cond$ of  $\Delta \mathbf{p}$ at $x^*$ is defined as follows :
	
	\begin{align}
	\nonumber
	Cond(x^*) &=
	\frac{||J(\mathbf{x}^{*})||}{|	\Delta \mathbf{p}(\mathbf{x}^{*})|/||\mathbf{x}^{*}||}\\
	&=\frac{\sqrt{\sum_{i=1}^{n}(\mathbf{x}_{i}^{*})^{2}}\sqrt{\sum_{i=1}^{n}(	 \frac{\partial\Delta\mathbf{p}}{\partial \mathbf{x}_{i}})^{2}(\mathbf{x}_{1}^{*},...,\mathbf{x}_{n}^{*})}}{|	\Delta \mathbf{p}(\mathbf{x}_{1}^{*},...,\mathbf{x}_{n}^{*})|}
	\label{eq:cond_num}
	\end{align}
	where $n$ denotes the number of the parameters. In this work, $n$ is equal to 8, and we adopt the $l_2$ norm in the equation. To analyze the distribution of condition number, we set a reasonable range for $\mathbf{x}$ according to the augmentation.
	
	 The right bar of Fig.~\ref{fig:cond_num} in the main paper denotes the probability value of a point in the coordinates, where the point x is sampled randomly for each parameter in a possible range. Fig.~\ref{fig:cond_numd} depicts a condition number changes in a subspace of the transformation range. The transformation ratio is chosen in the same increasing direction on eight parameters from non-transformation to the possible values

	\section{Implementation Details}
	\subsection{Rotation and Scale Estimation}
	Scale and rotation estimation takes a similar form as the one in translation estimation. Unlike the translation estimation, we do not crop the feature map in the neck of HDN. The feature map sizes of the output template and search image from the backbone are the same. 
	To facilitate the cross-correlation in Siamese Networks, it is necessary to extend the search map. Since translating the warped image is circular, we extend $\frac{n}{2}$ circularly as padding both on the top and bottom. In the horizontal direction, we simply adopt zero paddings.
	\begin{figure}[htbp]
		\centering 
		\includegraphics[scale=0.262]{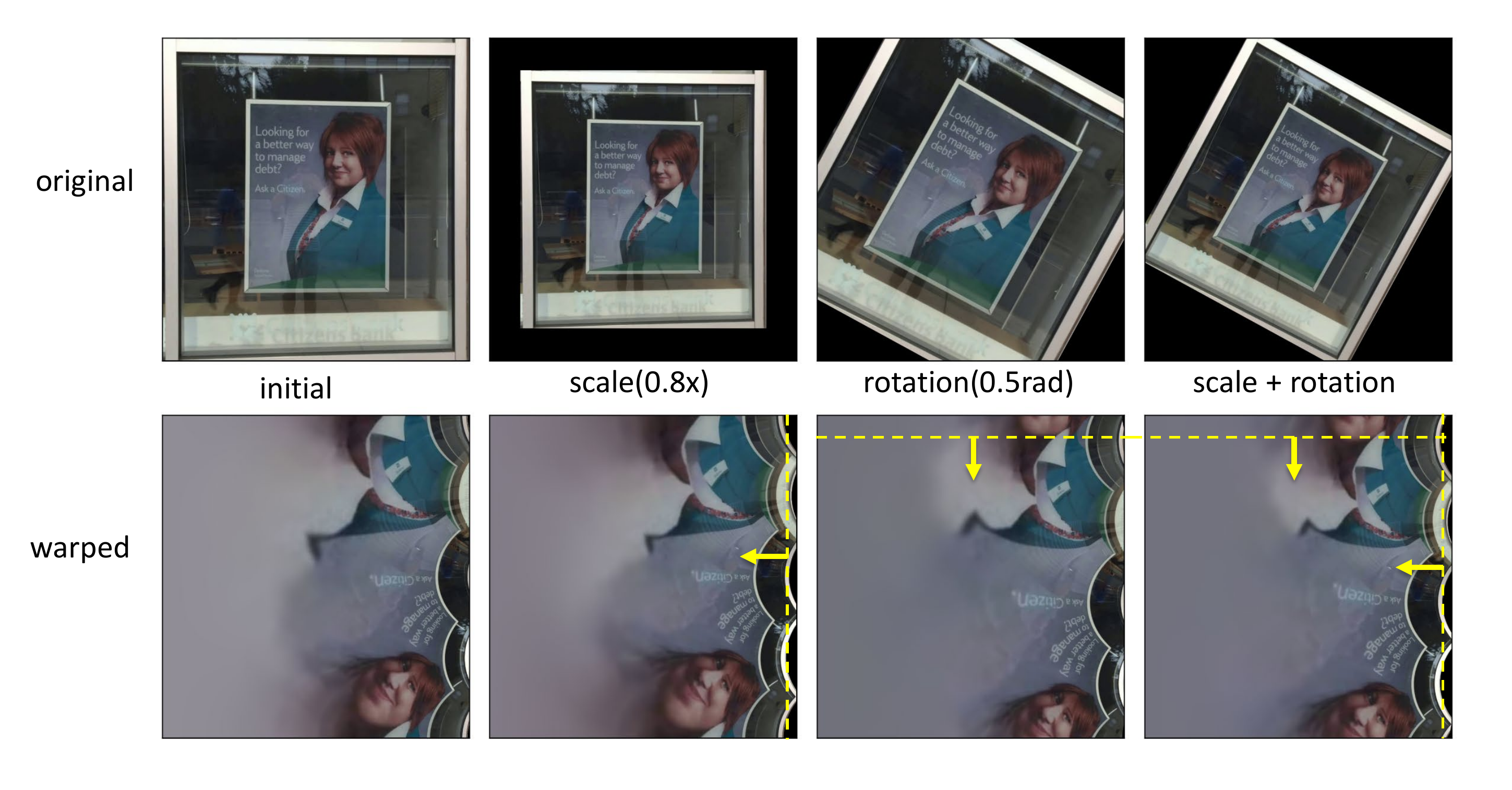}
		\caption{\footnotesize{Comparison on rotation and scale warping. The first row shows that the rotation and scale transformations are added to the initial image in Cartesian coordinates. The second row represents their corresponding warped images. It can be seen that the rotation and scale are merely the translation in the warped images.}}
	
		\label{fig:lp_illustration}
	\end{figure}

	The classification label map for the rotation and scale in training does not follow the translation estimating process.  In the log-polar coordinates, it is hard to define the probability map. As illustrated in Fig.~\ref{fig:lp_illustration}, every point shifting is the same within the image converted by RSEW. Due to the log operation in RSEW, most of the region in the warped image represents the foreground while the background only accounts for a small part. As a result, there are fewer hard negative samples, and we do not need to balance between hard and easy samples. Therefore, we use the same classification label as SiamBAN's \cite{SiamBAN}.

	\begin{figure*}[h]
		\centering 
		\includegraphics[scale=0.53]{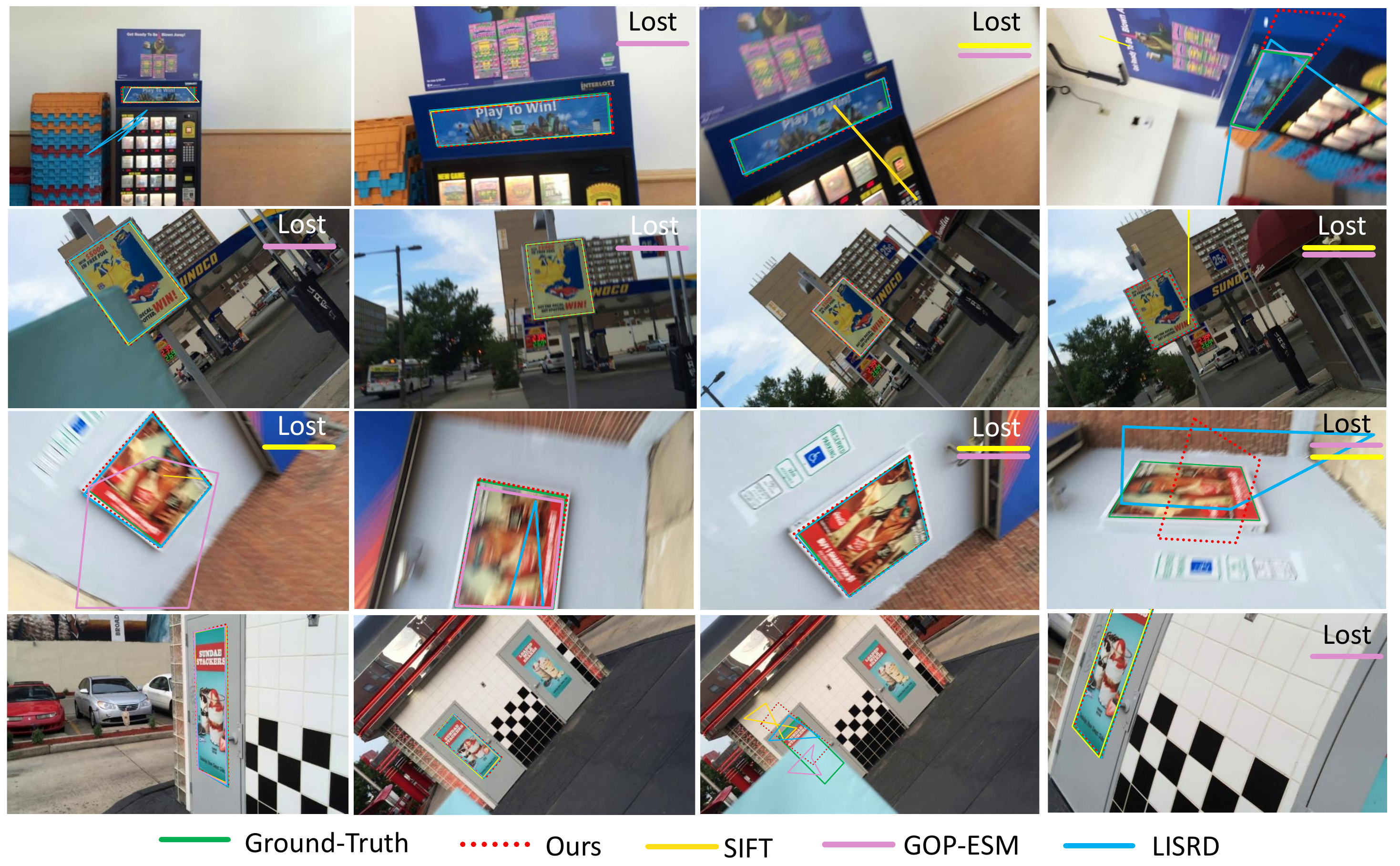}
		
		\caption{ A comparison of our approach with state-of-the-art trackers.}
		\label{fig:results_plot_all}
	\end{figure*}

	\subsection{Training}
	
	The backbone of the first stage network is a ResNet50~\cite{ResNet} pre-trained on ImageNet with reduced stride, where we choose C3, C4, and C5 layers features as our input. As in~\cite{SiamBAN}, the stride in conv4 and conv5 blocks are set to one, and the atrous rate is set to two and four, respectively. The second stage backbone is a regular ResNet34 pre-trained model, we change its input into two channels. ResNet50 is unchanged during training while the ResNet34 is trained with other modules.
	
	We trained the whole network on GOT-10k~\cite{GOT10k} and  COCO14~\cite{COCO} for 30 epochs with 1 epoch warming up. Moreover, we adopt Adam~\cite{adam} as the optimizer, where the batch size is set to 32$\times$4. Our model is trained in an end-to-end manner for 30 epochs. The learning rate of the first epoch for warming up is set from 0.001 to 0.005, and the learning rate for the rest of the epochs is from 0.0012 to $2\times 10^{-6}$. An exponential learning scheduler is adopted with 
	the multiplicative factor of learning rate decay $\beta=0.8$.
	
	%	\textbf{Tracking}.
	\begin{figure*}[htbp]
		\centering
		\subfigure{
			\begin{minipage}[t]{0.25\linewidth}
				\centering
				\includegraphics[width=1\linewidth]{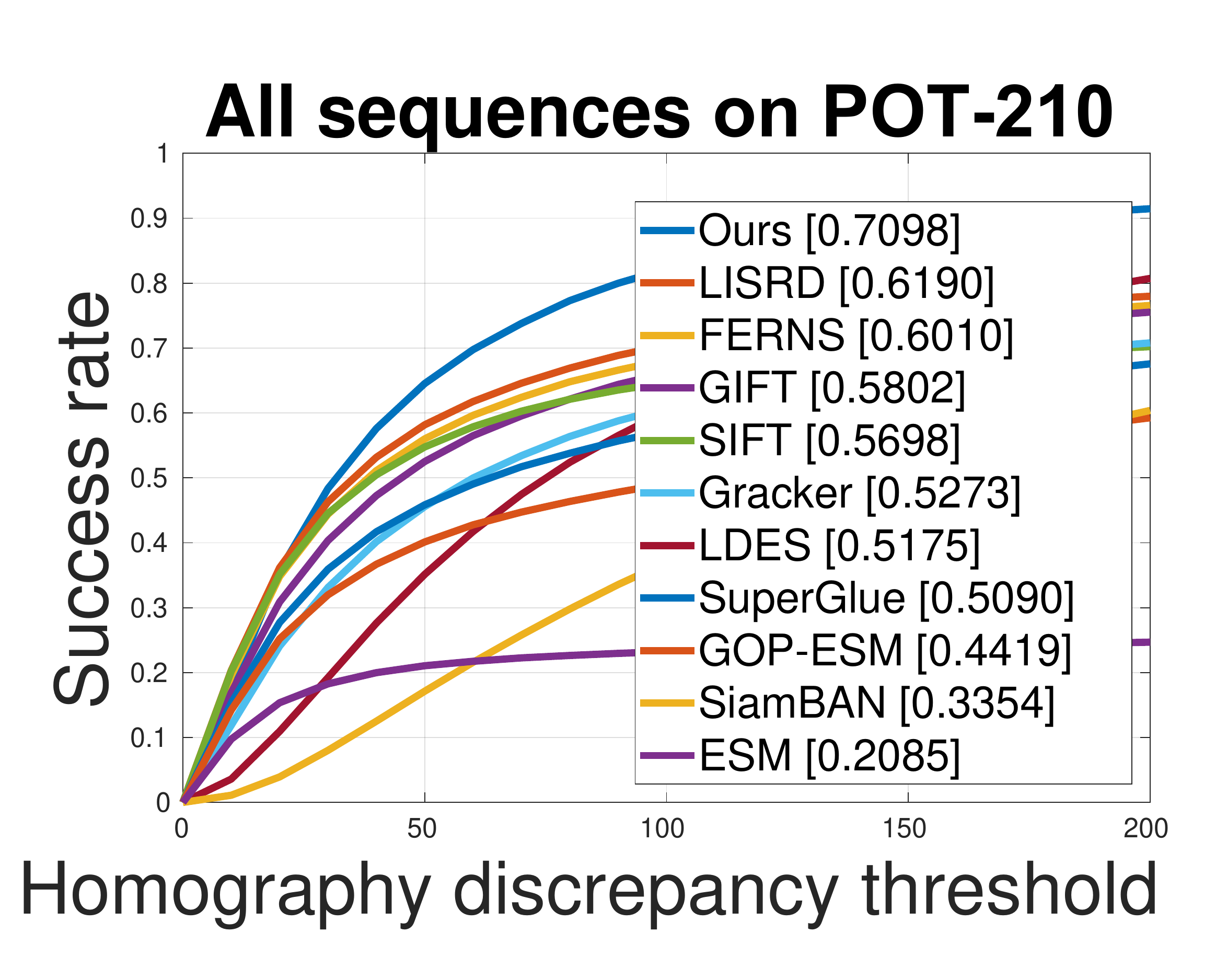}
				%\caption{fig2}
			\end{minipage}
		}%
		\subfigure{
			\begin{minipage}[t]{0.25\linewidth}
				\centering
				\includegraphics[width=1\linewidth]{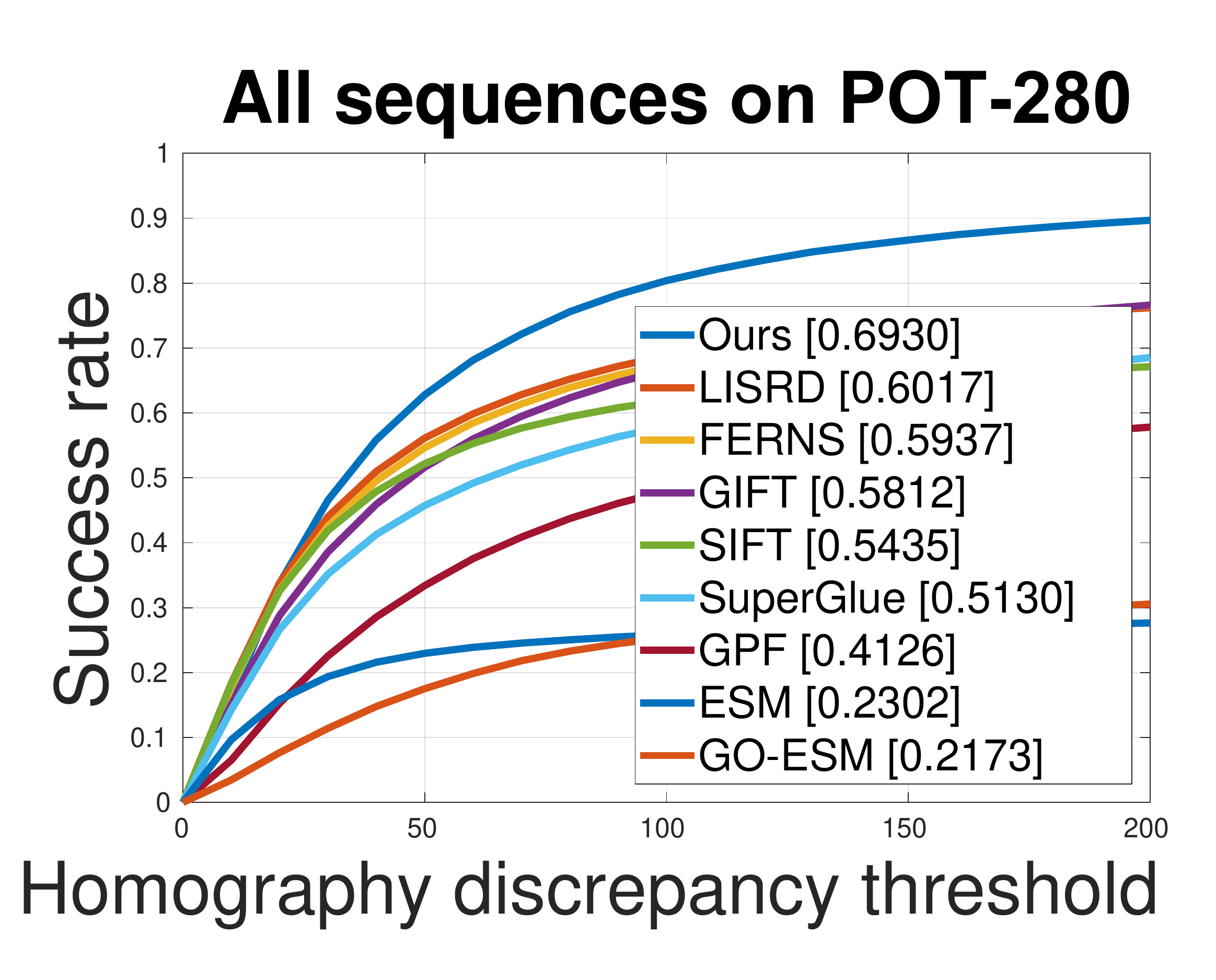}
				%\caption{fig2}
			\end{minipage}
		}%
		\subfigure{
			\begin{minipage}[t]{0.25\linewidth}
				\centering
				\includegraphics[width=1\linewidth]{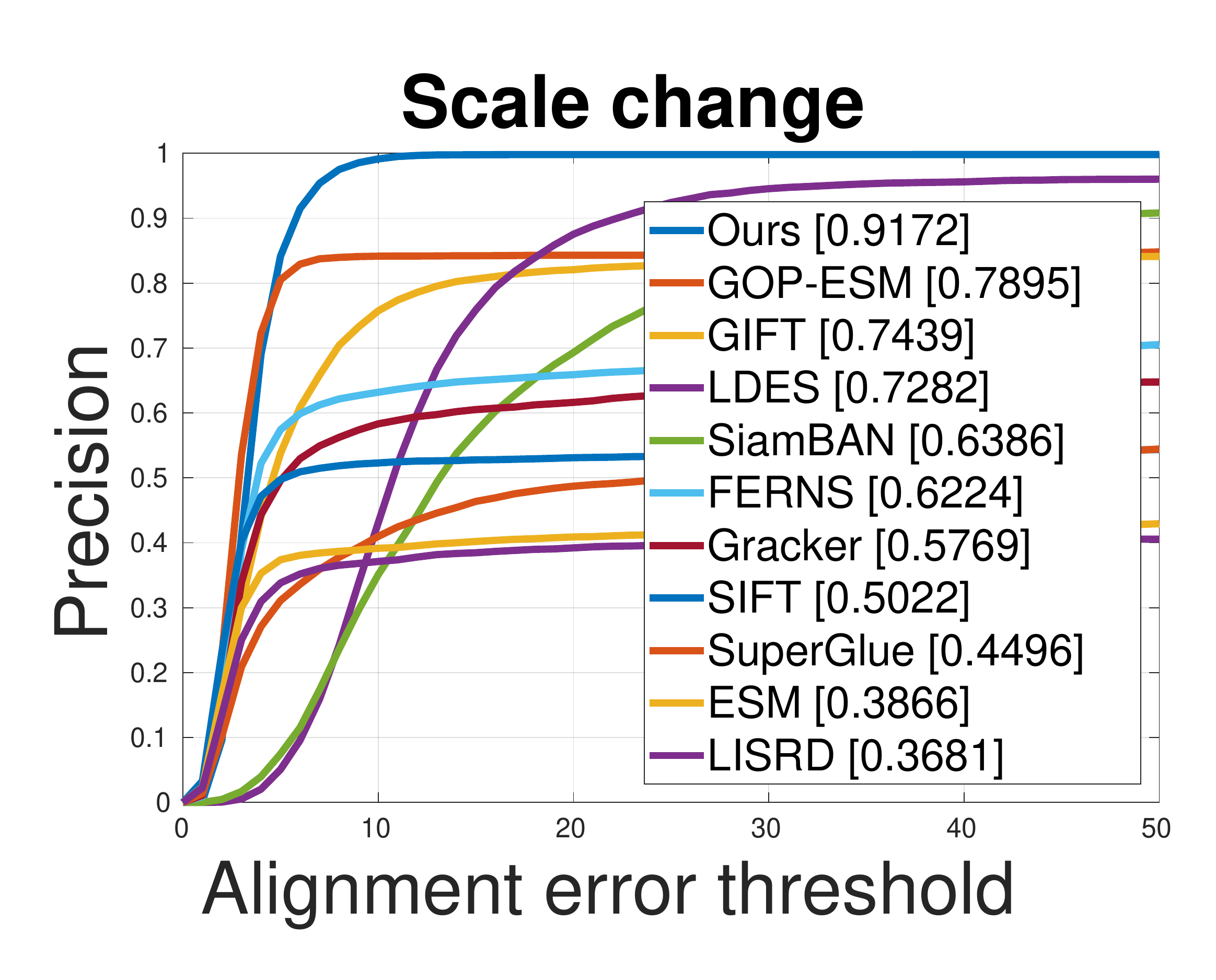}
				%\caption{fig1}
			\end{minipage}%
		}%
		\subfigure{
			\begin{minipage}[t]{0.25\linewidth}
				\centering
				\includegraphics[width=1\linewidth]{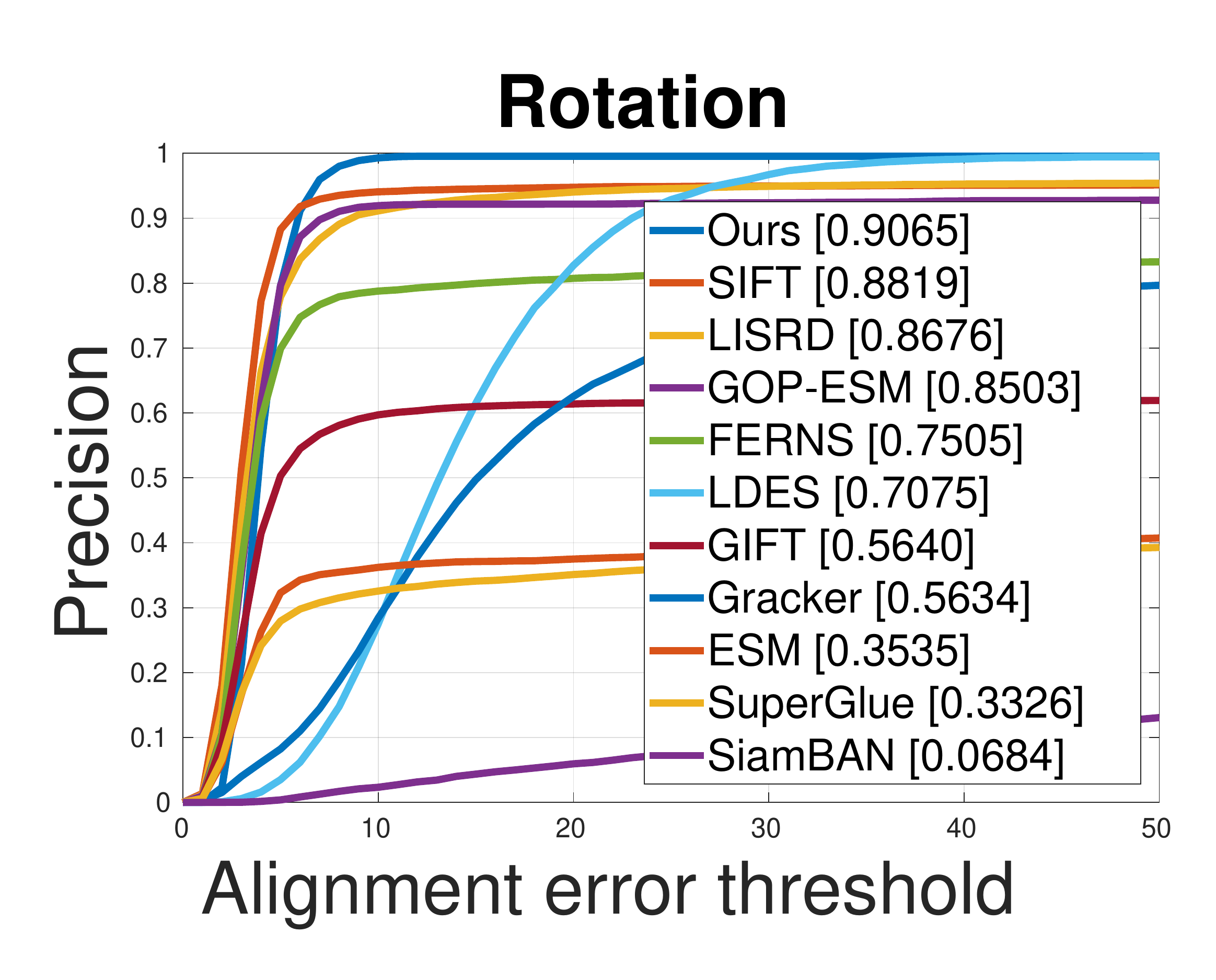}
				%\caption{fig2}
			\end{minipage}
		}%

		\subfigure{
			\begin{minipage}[t]{0.25\linewidth}
				\centering
				\includegraphics[width=1\linewidth]{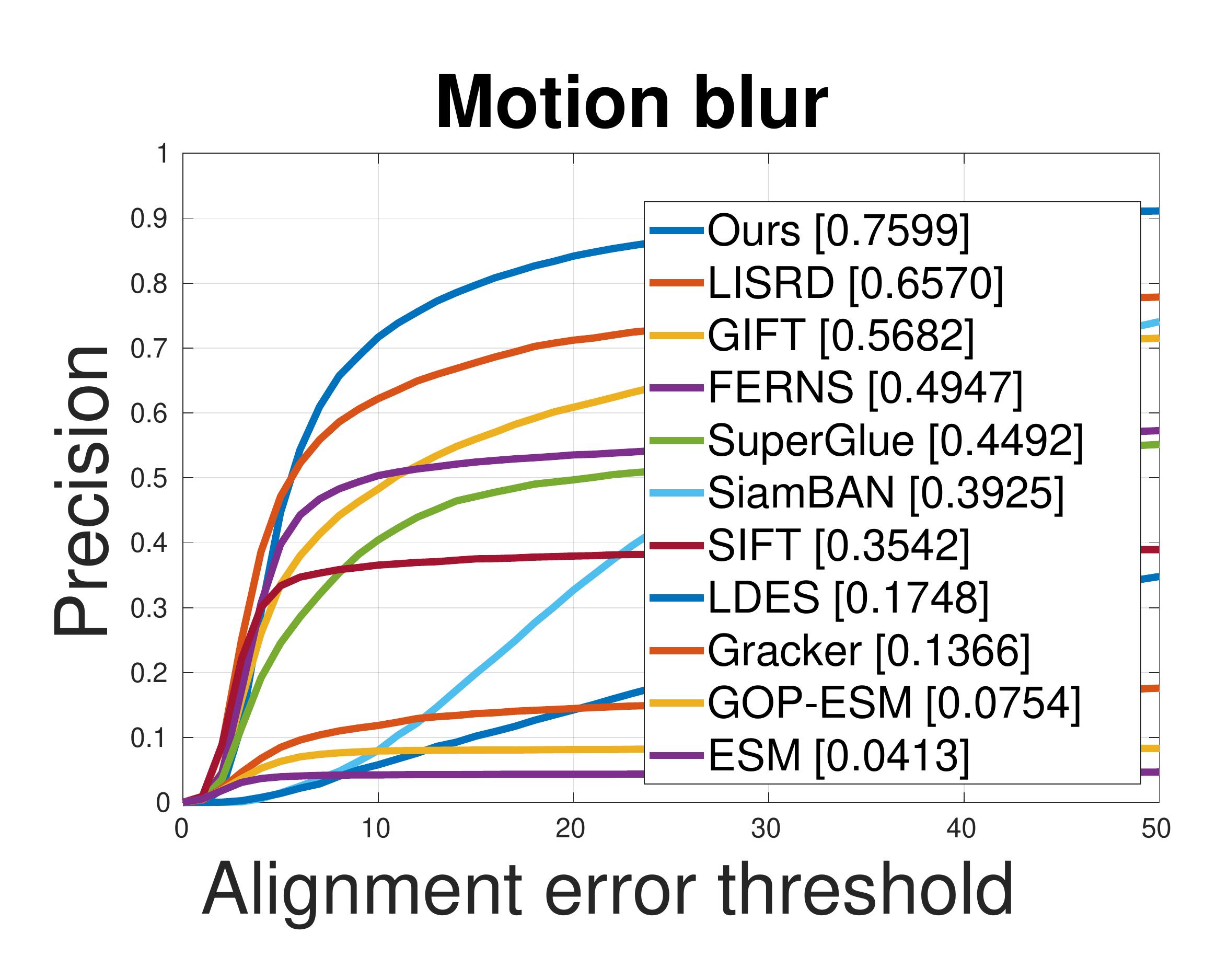}
				%\caption{fig2}
			\end{minipage}
		}%
		\subfigure{
			\begin{minipage}[t]{0.25\linewidth}
				\centering
				\includegraphics[width=1\linewidth]{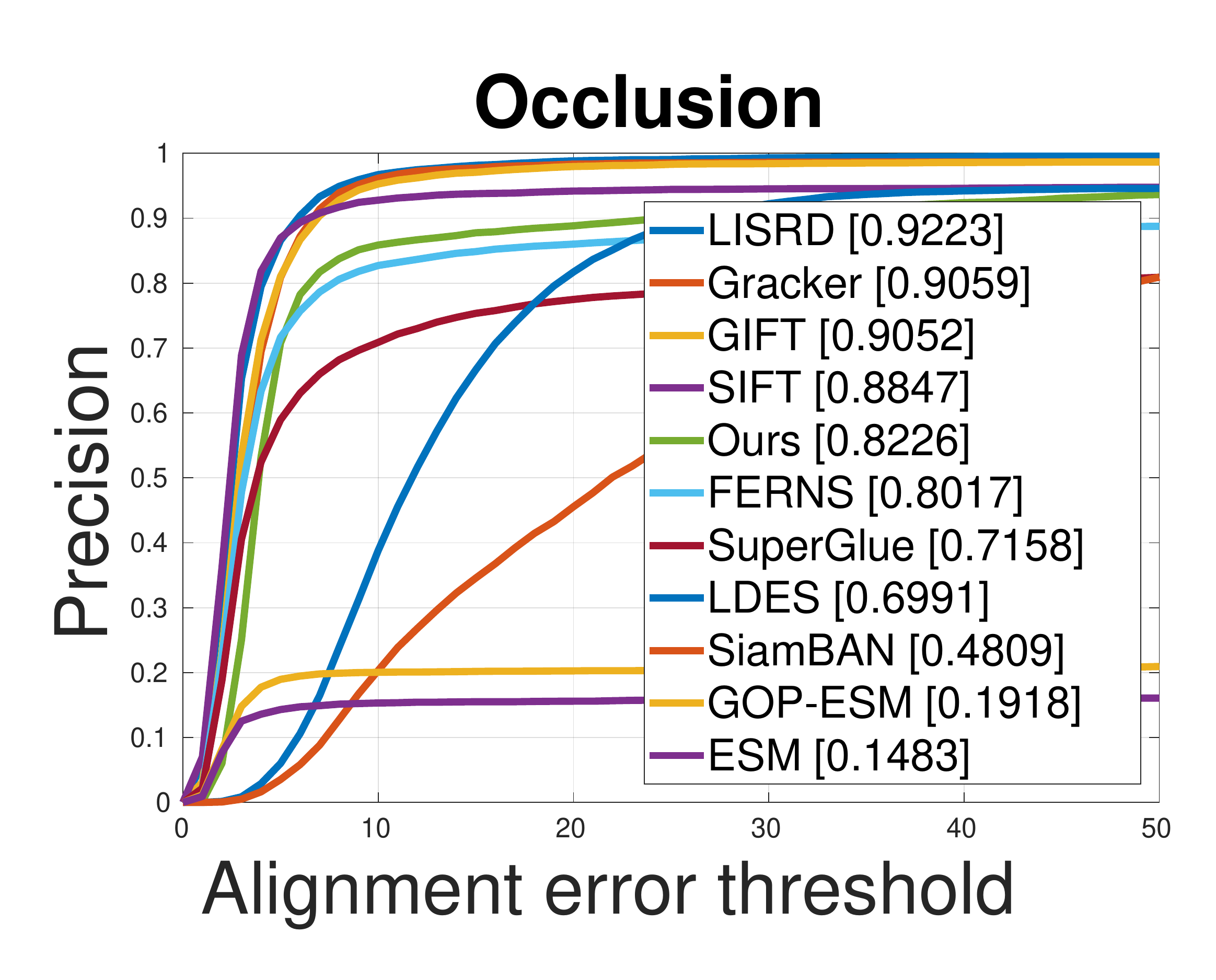}
				%\caption{fig2}
			\end{minipage}
		}%	
		\subfigure{
			\begin{minipage}[t]{0.25\linewidth}
				\centering
				\includegraphics[width=1\linewidth]{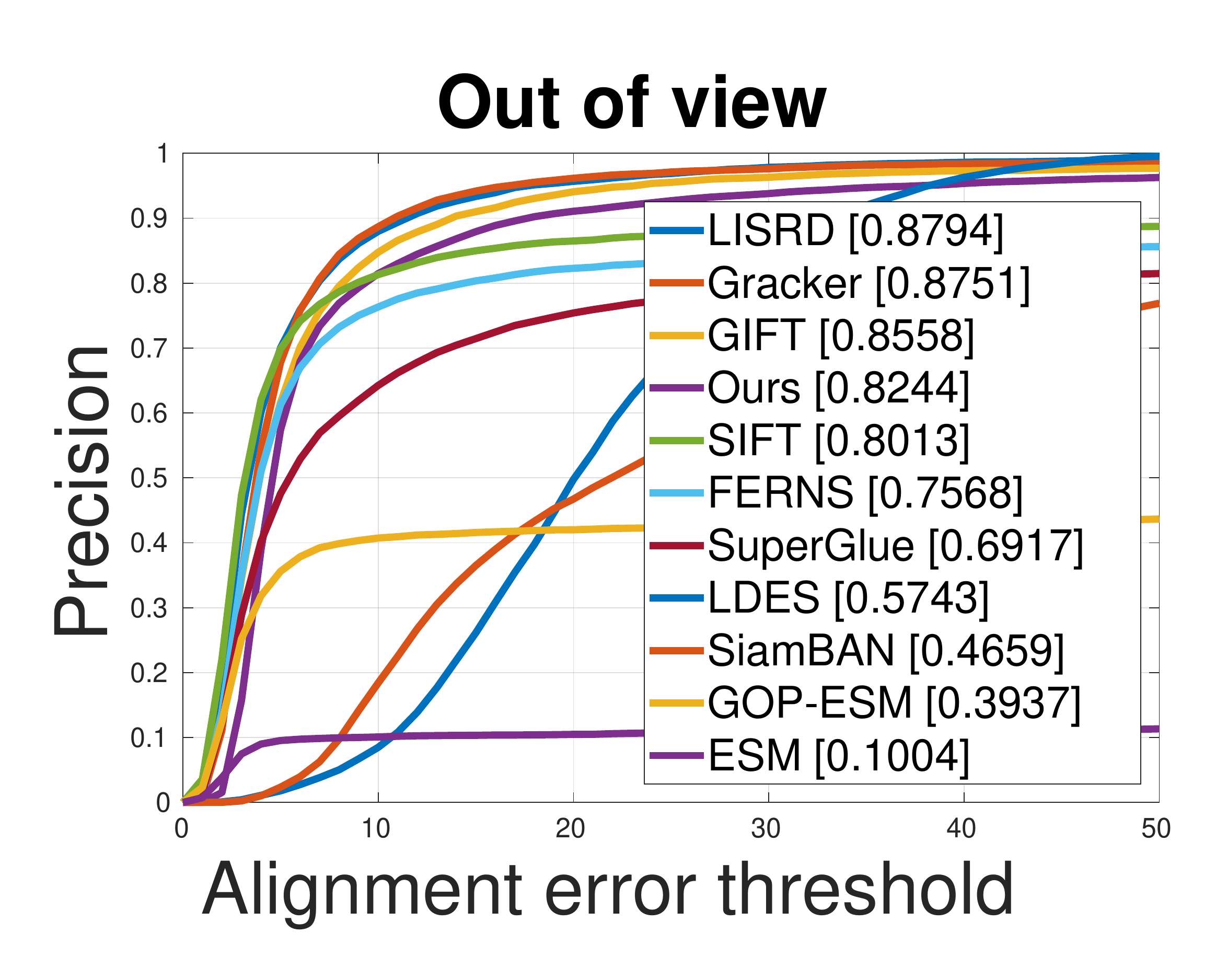}
				%\caption{fig1}
			\end{minipage}%
		}%
		\subfigure{
			\begin{minipage}[t]{0.25\linewidth}
				\centering
				\includegraphics[width=1\linewidth]{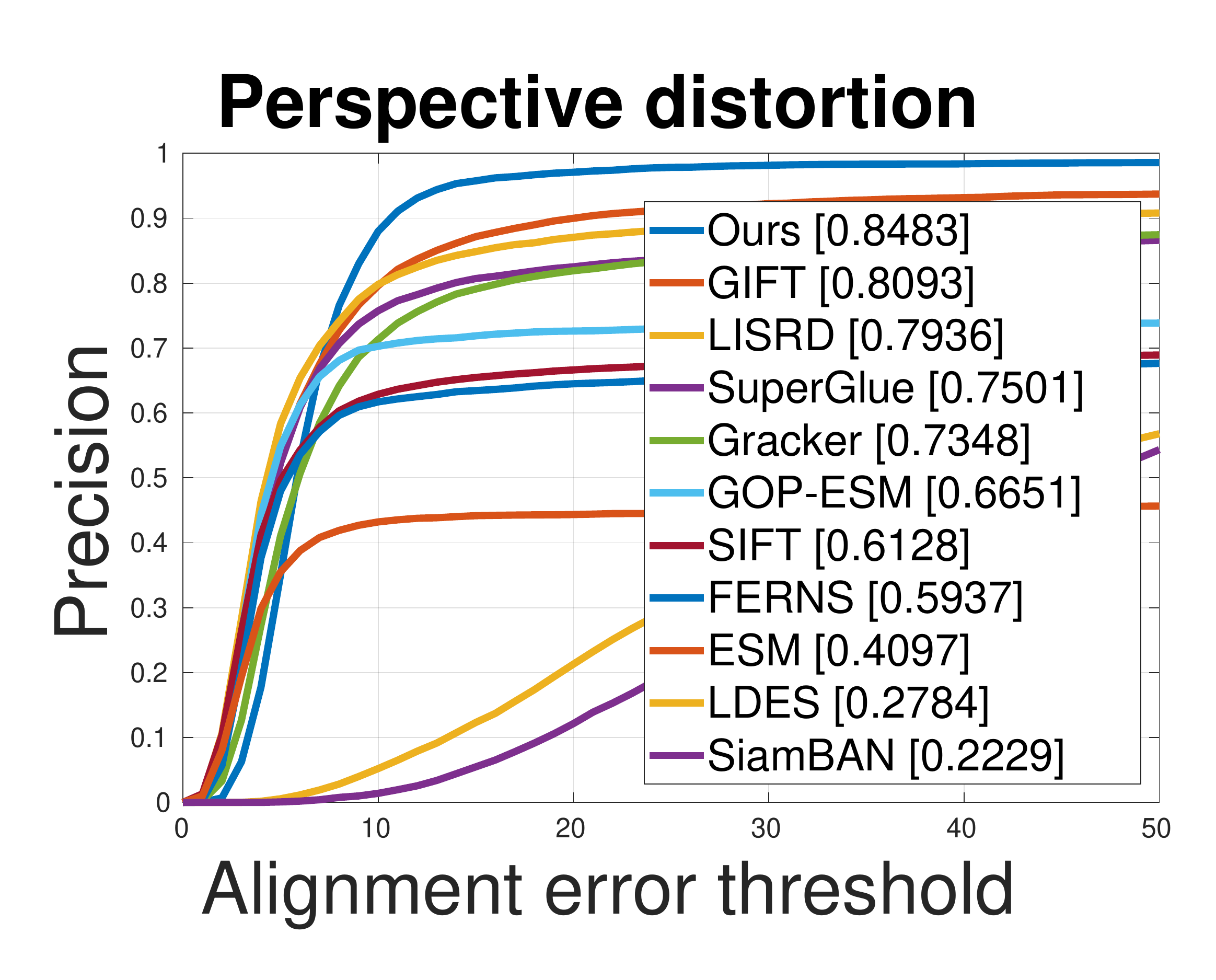}
				%\caption{fig2}
			\end{minipage}
		}%

		\subfigure{
			\begin{minipage}[t]{0.25\linewidth}
				\centering
				\includegraphics[width=1\linewidth]{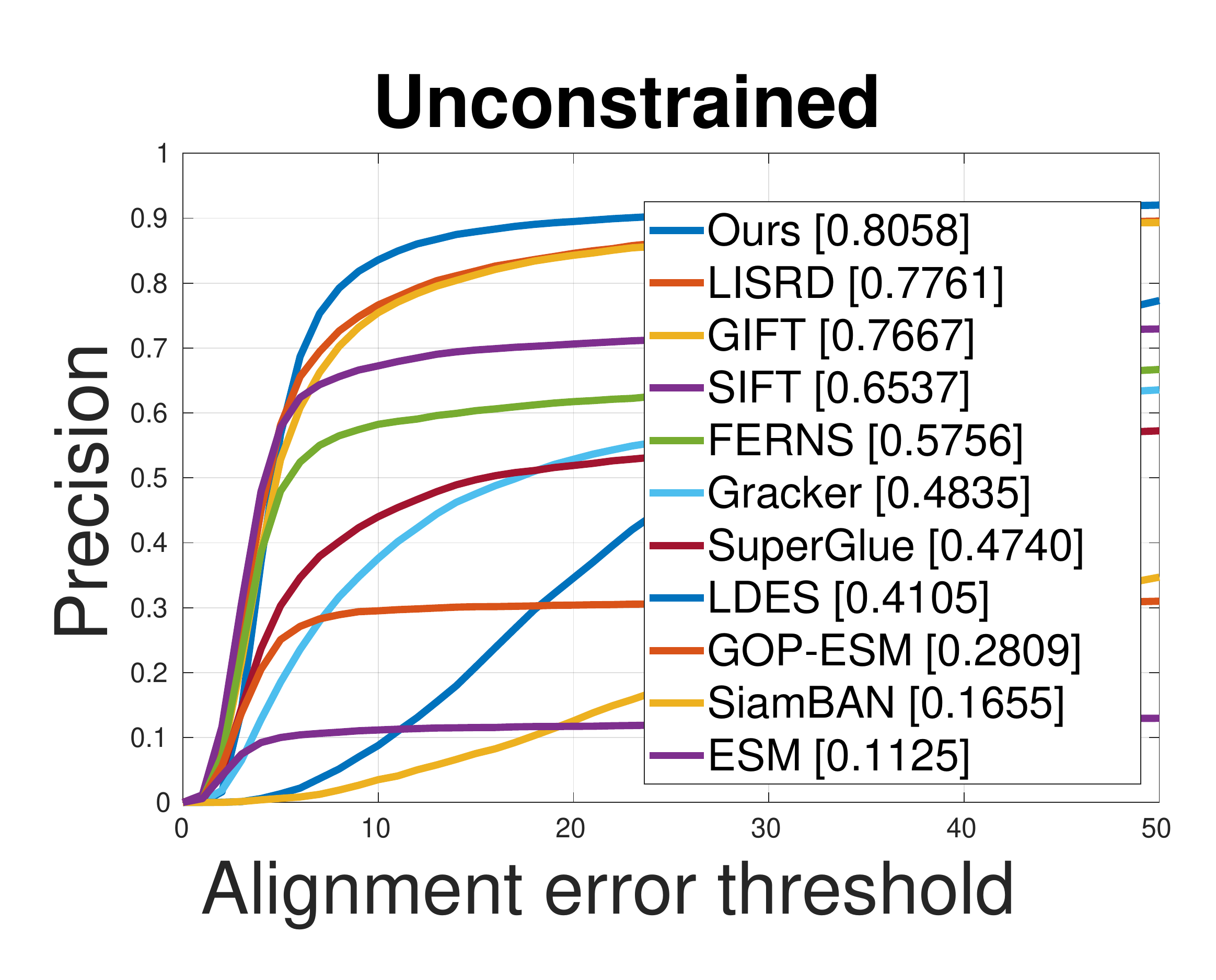}
				%\caption{fig2}
			\end{minipage}
		}%
		\subfigure{
			\begin{minipage}[t]{0.25\linewidth}
				\centering
				\includegraphics[width=1\linewidth]{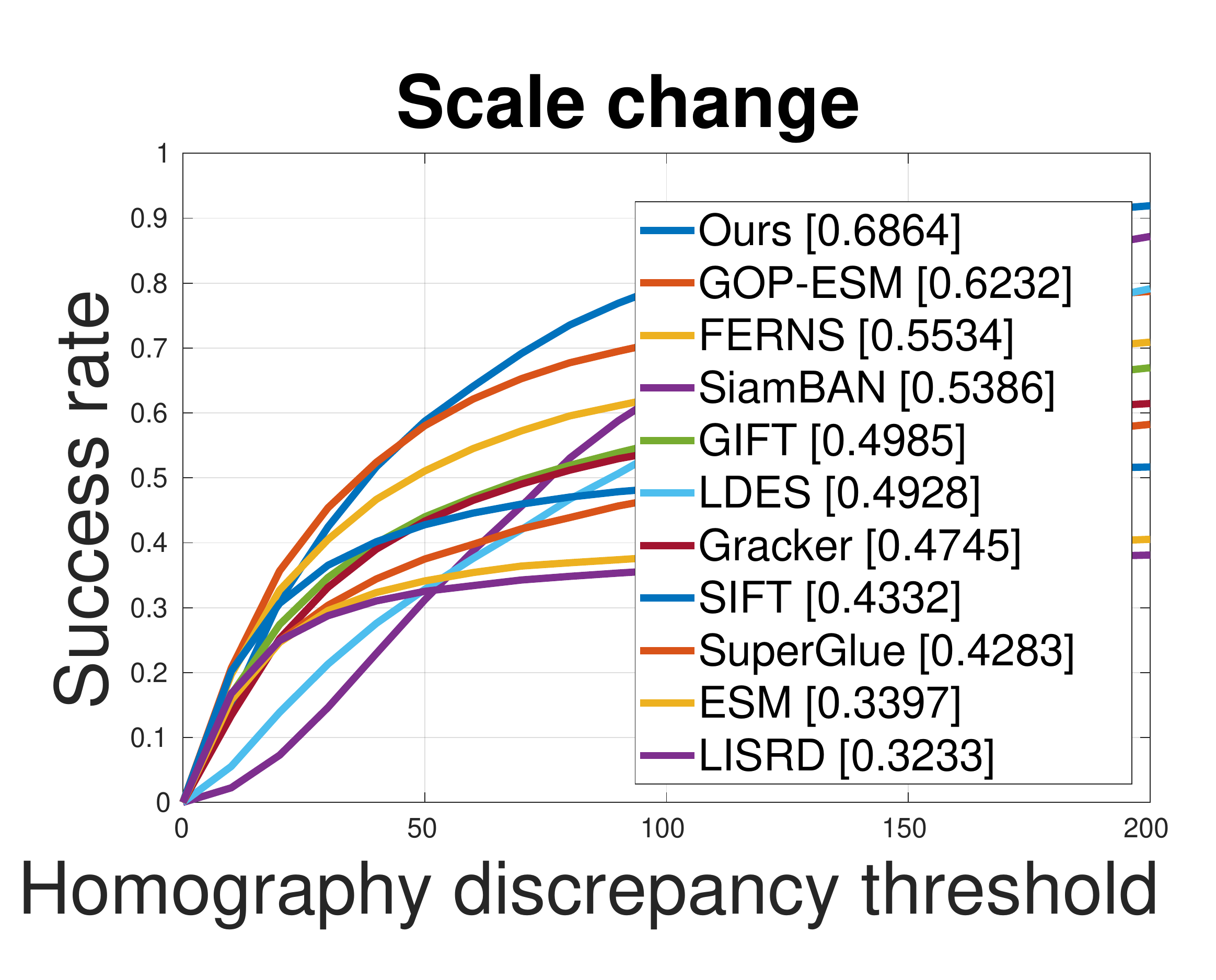}
				%\caption{fig1}
			\end{minipage}%
		}%
		\subfigure{
			\begin{minipage}[t]{0.25\linewidth}
				\centering
				\includegraphics[width=1\linewidth]{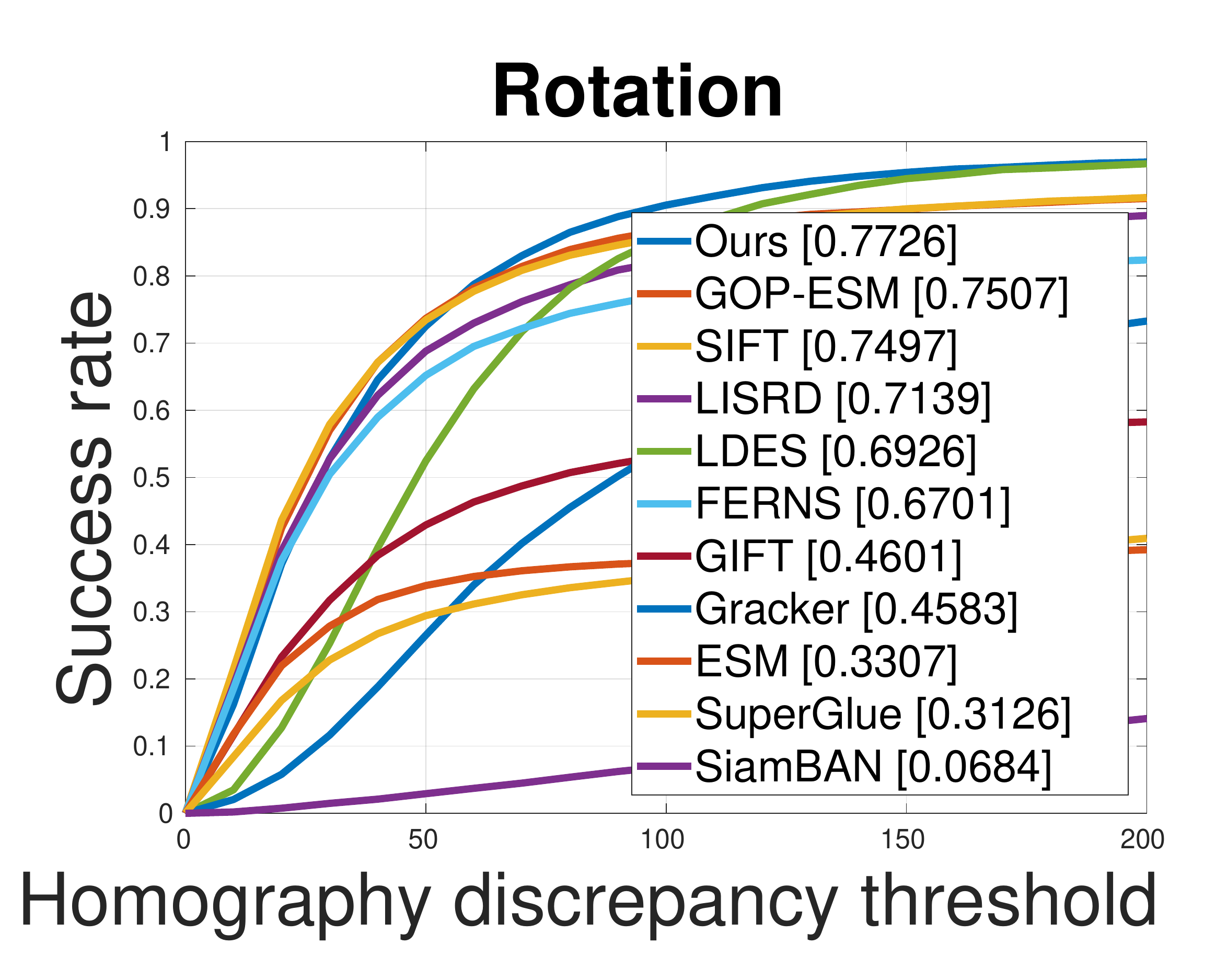}
				%\caption{fig2}
			\end{minipage}
		}%
		\subfigure{
			\begin{minipage}[t]{0.25\linewidth}
				\centering
				\includegraphics[width=1\linewidth]{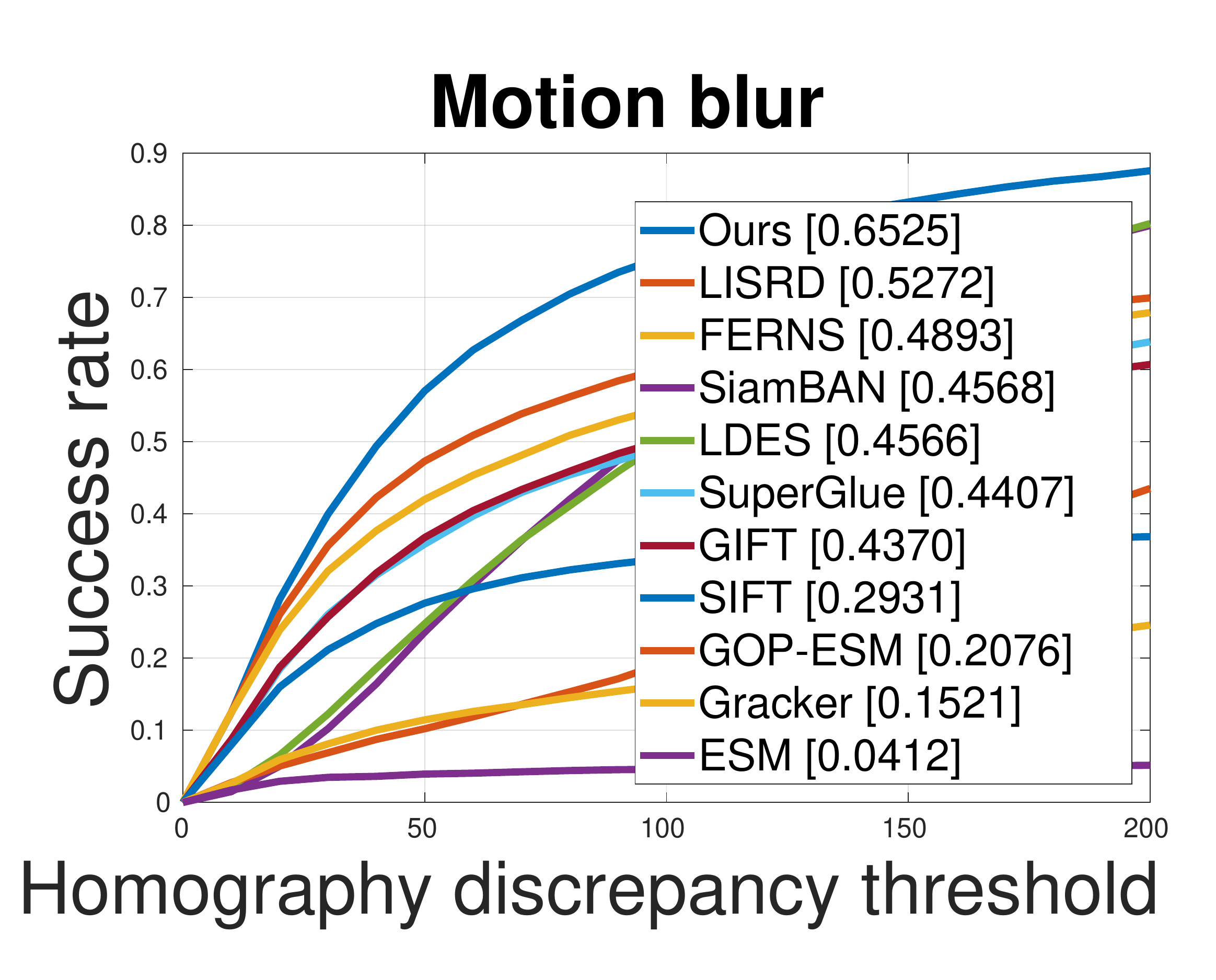}
				%\caption{fig2}
			\end{minipage}
		}%
		
		\subfigure{
			\begin{minipage}[t]{0.25\linewidth}
				\centering
				\includegraphics[width=1\linewidth]{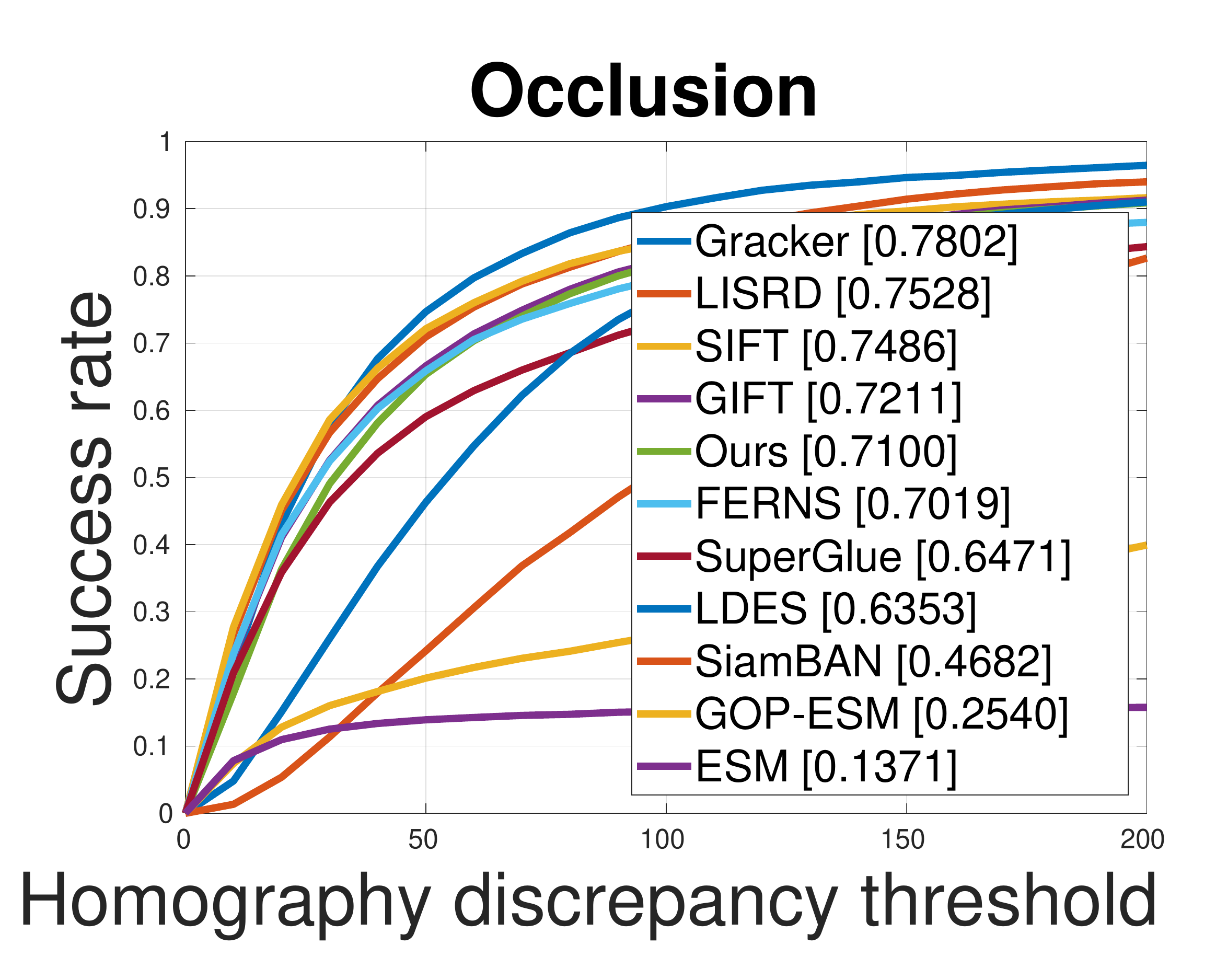}
				%\caption{fig2}
			\end{minipage}
		}%
		\subfigure{
			\begin{minipage}[t]{0.25\linewidth}
				\centering
				\includegraphics[width=1\linewidth]{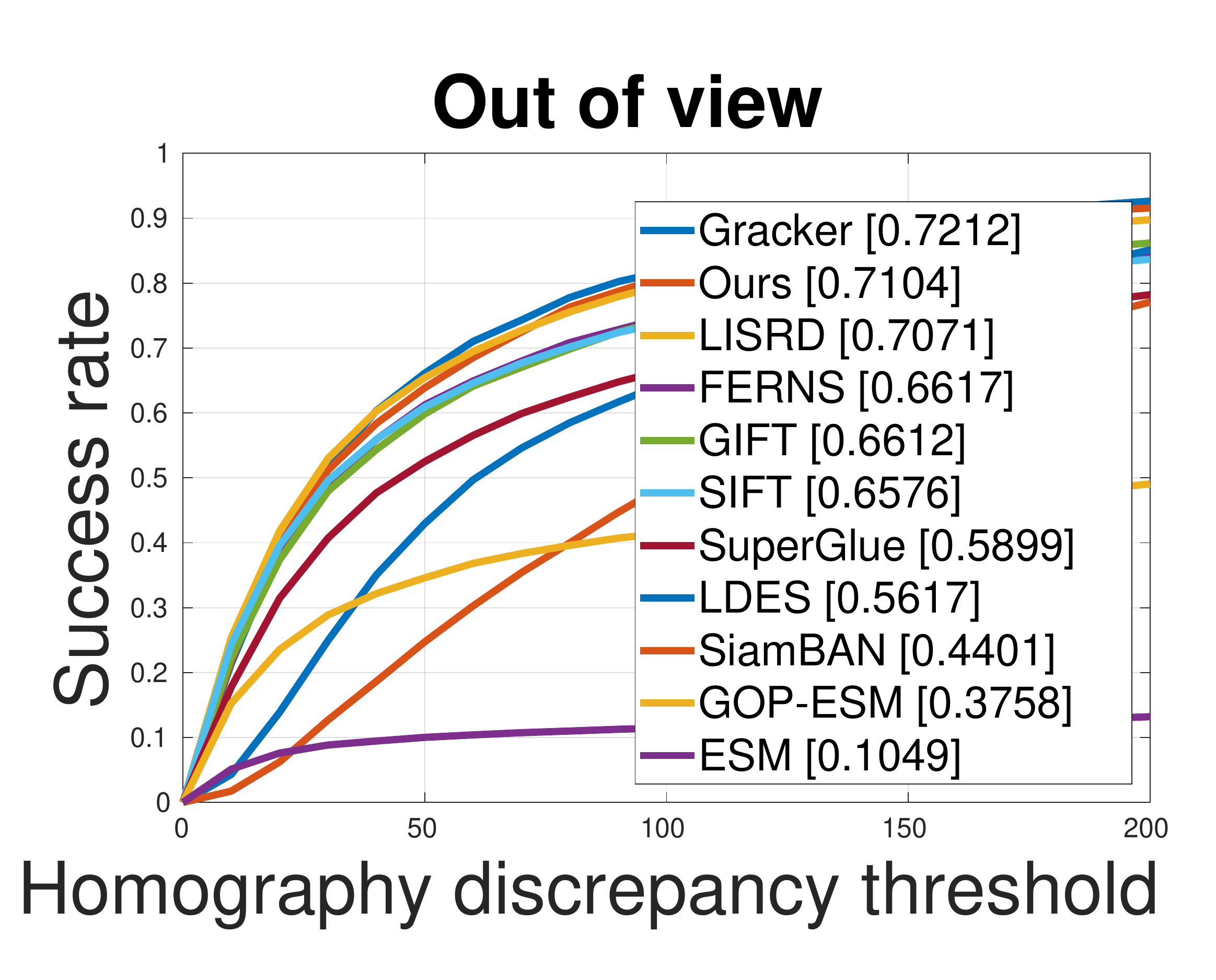}
				%\caption{fig1}
			\end{minipage}%
		}%
		\subfigure{
			\begin{minipage}[t]{0.25\linewidth}
				\centering
				\includegraphics[width=1\linewidth]{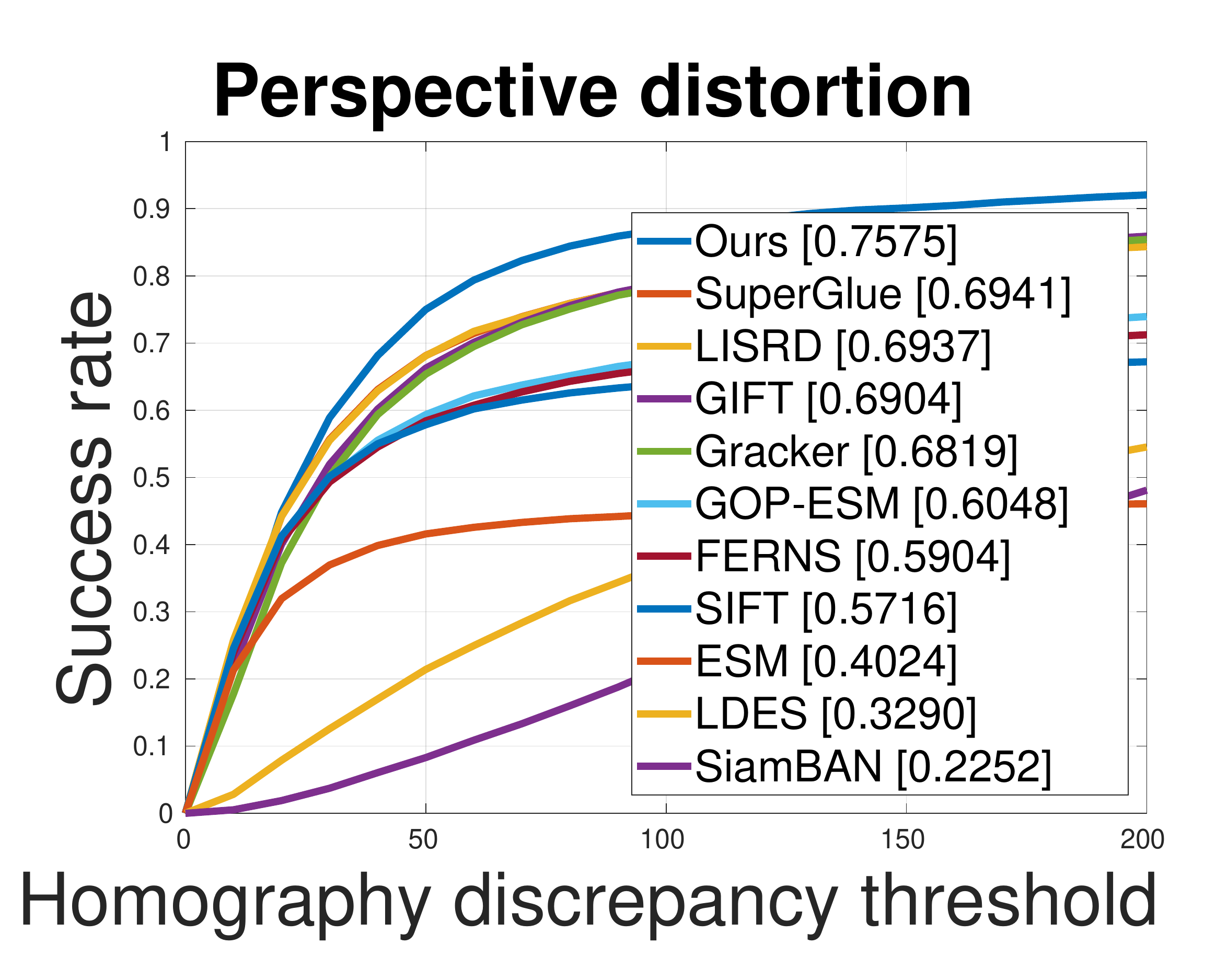}
				%\caption{fig2}
			\end{minipage}
		}%
		\subfigure{
			\begin{minipage}[t]{0.25\linewidth}
				\centering
				\includegraphics[width=1\linewidth]{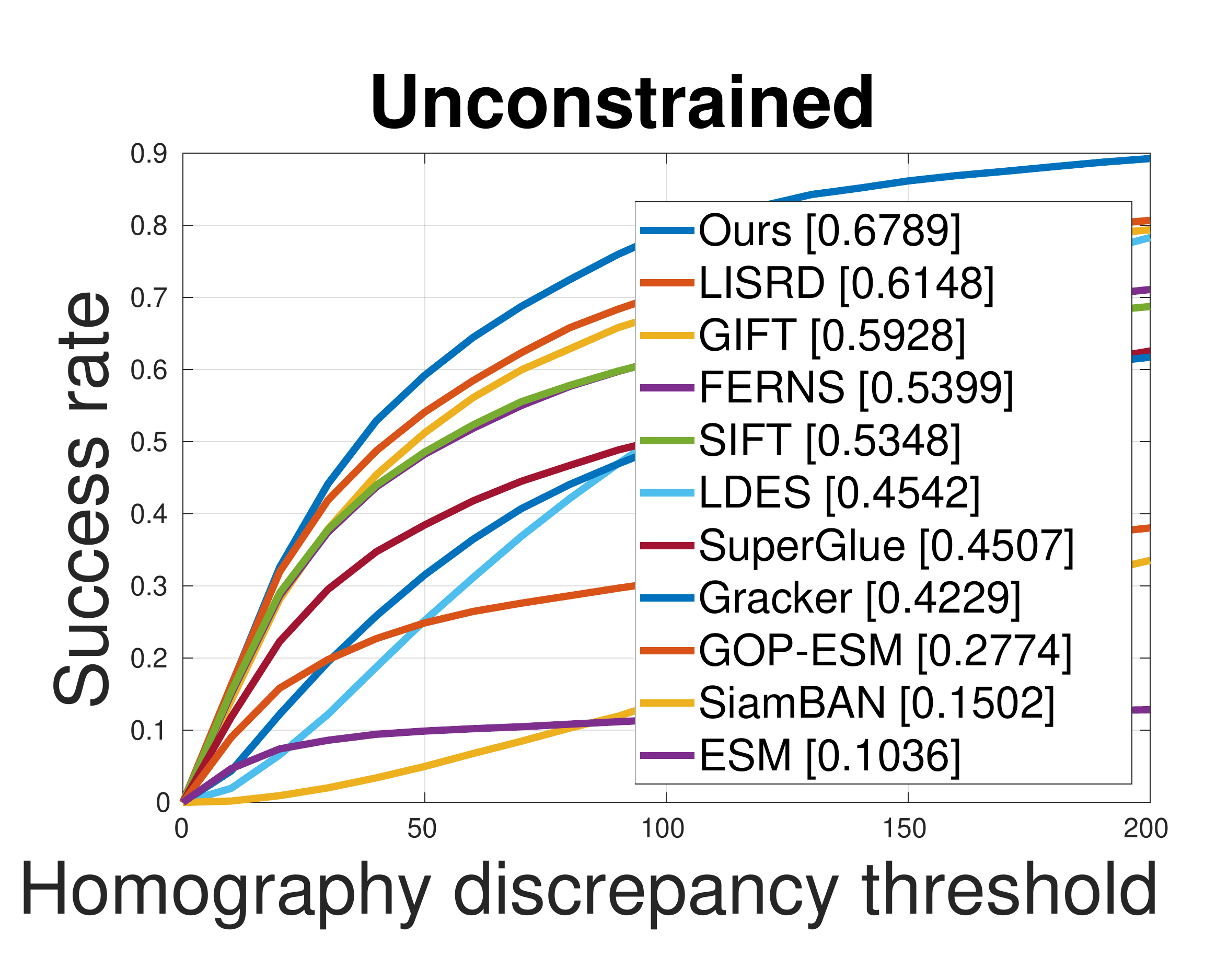}
				%\caption{fig2}
			\end{minipage}
		}%
		
		\centering
		\caption{Comparisons on POT \cite{POT} with seven challenging factors. The first two subfigures depict the Homography Success Rate on POT-210 and POT-280, respectively. Other subfigures give the Precision and Homography Success rates on seven challenging scenarios.  }
		%		left figure and middle figure are the Precision and Homography Success Rate respectively with a different threshold. Legends show the Average precision and Homography Success Rate. The third radar subfigure compare the trackers' Average Precision on 7 challenging factors}
		\label{fig:pot_compare_more}
	\end{figure*}
	\begin{table*}[t]
		\centering
		\resizebox{0.75\linewidth}{!}{
			\begin{tabular}{p{4.89em}|lllllll}
				\hline
				\textbf{tracker} & \multicolumn{1}{p{5.28em}}{\textbf{avg \newline{}Prec}} & \multicolumn{1}{p{5.28em}}{\textbf{avg\newline{}HSR}} & \multicolumn{1}{p{3.72em}}{\textbf{Prec \newline{}(e$\leq$ 5)}} & \multicolumn{1}{p{3.835em}}{\textbf{Prec \newline{}(e$\leq$ 10)}} & \multicolumn{1}{p{3.39em}}{\textbf{Prec \newline{}(e$\leq$ 20)}} & \multicolumn{1}{p{3.39em}}{\textbf{avg\newline{}CP}} & \multicolumn{1}{p{3.665em}}{\textbf{avg \newline{}SR}} \\
                 \hline
                    SIFT  & 0.670  & 0.570  & \textcolor[rgb]{ 1,  0,  0}{\textbf{0.622 }} & 0.695  & \textcolor[rgb]{ .329,  .51,  .208}{\textbf{0.719 }} & 0.698  & 0.701  \\
                    SURF  & 0.657  & 0.536  & 0.543  & 0.663  & 0.711  & 0.689  & 0.705  \\
                    LISRD & \textcolor[rgb]{ .184,  .459,  .71}{\textbf{0.752 }} & \textcolor[rgb]{ .184,  .459,  .71}{\textbf{0.619 }} & \textcolor[rgb]{ .184,  .459,  .71}{\textbf{0.617 }} & \textcolor[rgb]{ .184,  .459,  .71}{\textbf{0.759 }} & \textcolor[rgb]{ .184,  .459,  .71}{\textbf{0.815 }} & \textcolor[rgb]{ .184,  .459,  .71}{\textbf{0.788 }} & \textcolor[rgb]{ .184,  .459,  .71}{\textbf{0.805 }} \\
                    GIFT  & \textcolor[rgb]{ .329,  .51,  .208}{\textbf{0.745}}  & 0.580  & 0.551  & \textcolor[rgb]{ .329,  .51,  .208}{\textbf{0.741 }} & \textcolor[rgb]{ .184,  .459,  .71}{\textbf{0.815 }} & \textcolor[rgb]{ .329,  .51,  .208}{\textbf{0.785 }} & \textcolor[rgb]{ .329,  .51,  .208}{\textbf{0.801 }} \\
                    SuperGlue & 0.552  & 0.509  & 0.389  & 0.526  & 0.601  & 0.592  & 0.659  \\
                    Gracker & 0.611  & 0.527  & 0.392  & 0.560  & 0.668  & 0.684  & 0.716  \\
                    ESM   & 0.222  & 0.209  & 0.204  & 0.227  & 0.235  & 0.242  & 0.261  \\
                    FERNS & 0.657  & \textcolor[rgb]{ .329,  .51,  .208}{\textbf{0.601}}  & 0.565  & 0.673  & 0.706  & 0.686  & 0.724  \\
                    SCV   & 0.250  & 0.233  & 0.228  & 0.257  & 0.265  & 0.274  & 0.287  \\
                    GOP-ESM & 0.464  & 0.442  & 0.430  & 0.492  & 0.499  & 0.484  & 0.497  \\
                    SiamBAN & 0.348  & 0.335  & 0.022  & 0.127  & 0.321  & 0.650  & 0.693  \\
                    LDES  & 0.510  & 0.518  & 0.028  & 0.196  & 0.531  & 0.617  & 0.737  \\
                    Ocean & 0.255  & 0.261  & 0.011  & 0.089  & 0.241  & 0.464  & 0.521  \\
                    SOSNet & 0.658  & 0.537  & 0.566  & 0.674  & 0.711  & 0.688  & 0.693  \\
                    LIFT  & 0.580  & 0.485  & 0.503  & 0.593  & 0.625  & 0.608  & 0.623  \\
                    MatchNet & 0.595  & 0.491  & 0.521  & 0.612  & 0.642  & 0.620  & 0.621  \\
                    SOL   & 0.511  & 0.448  & 0.416  & 0.511  & 0.550  & 0.540  & 0.592  \\
                    GPF   & 0.510  & 0.405  & 0.287  & 0.489  & 0.569  & 0.555  & 0.589  \\
                    MCPF  & 0.181  & 0.285  & 0.004  & 0.034  & 0.139  & 0.412  & 0.537  \\
                    L1APG & 0.180  & 0.201  & 0.024  & 0.083  & 0.177  & 0.283  & 0.377  \\
                    CNN-GM & 0.228  & 0.190  & 0.079  & 0.198  & 0.259  & 0.278  & 0.345  \\
                    IVT   & 0.208  & 0.184  & 0.012  & 0.098  & 0.225  & 0.332  & 0.334  \\
                    PFNet & 0.203  & 0.168  & 0.113  & 0.182  & 0.222  & 0.247  & 0.317  \\
                    GO-ESM & 0.211  & 0.184  & 0.143  & 0.212  & 0.232  & 0.241  & 0.314  \\
                    SiamMask & 0.348  & 0.323  & 0.029  & 0.133  & 0.323  & 0.644  & 0.684  \\
                    SiamRPN & 0.354  & 0.330  & 0.021  & 0.127  & 0.331  & 0.641  & 0.691  \\
                    SiamRPN++ & 0.370  & 0.340  & 0.034  & 0.154  & 0.360  & 0.658  & 0.702  \\
                    ECO   & 0.360  & 0.378  & 0.025  & 0.128  & 0.326  & 0.548  & 0.681  \\
                    TransT & 0.464  & 0.381  & 0.090  & 0.276  & 0.486  & 0.764  & 0.774  \\
                    \hline
                    \textbf{Ours} & \textcolor[rgb]{ 1,  0,  0}{\textbf{0.841 }} & \textcolor[rgb]{ 1,  0,  0}{\textbf{0.710 }} & \textcolor[rgb]{ .329,  .51,  .208}{\textbf{0.611 }} & \textcolor[rgb]{ 1,  0,  0}{\textbf{0.870 }} & \textcolor[rgb]{ 1,  0,  0}{\textbf{0.928 }} & \textcolor[rgb]{ 1,  0,  0}{\textbf{0.886 }} & \textcolor[rgb]{ 1,  0,  0}{\textbf{0.904 }} \\
                    \hline
				
			\end{tabular}%
		}
		\caption{Results on POT-210. Top-3 results of each dimension are colored in red, blue and green, respectively.}
		\label{tab:POT_full_results}%
	\end{table*}%

	\begin{table*}[h]
		\centering
		\resizebox{\linewidth}{!}{
			\begin{tabular}{c|ccccccccccccccccc|cc}
				\hline
				{\textbf{Trackers}} & \textbf{SIFT} & \textbf{Gracker} & \textbf{TransT} & \textbf{GOP-ESM} & \textbf{GO-ESM} & \textbf{SOL} & \textbf{FERNS} & \textbf{Bit-Planes} & \textbf{MI} & \textbf{LSCV} & \textbf{SCV} & \textbf{ESM-RBF} & \textbf{CCRE} & \textbf{ESM-GB} & \textbf{NCC} & \textbf{NGF} & \textbf{ESM} & \textbf{Ours} \\
				\hline
				 \textbf{avg Prec} & 0.527 & \textcolor[rgb]{ .329,  .51,  .208}{\textbf{0.819}} & 0.367 & \textcolor[rgb]{ 0,  .439,  .753}{\textbf{0.873}} & 0.78  & 0.678 & 0.632 & 0.694 & 0.491 & 0.594 & 0.539 & 0.555 & 0.472 & 0.508 & 0.461 & 0.322 & 0.417 & \textcolor[rgb]{ 1,  0,  0}{\textbf{0.874}} \\
				\textbf{Prec(e$\leq$ 5)} & 0.4   & 0.671 & 0.047 & \textcolor[rgb]{ 1,  0,  0}{\textbf{0.868}} &  \textcolor[rgb]{ .329,  .51,  .208}{\textbf{0.715}} & 0.483 & 0.442 & 0.701 & 0.479 & 0.594 & 0.533 & 0.547 & 0.456 & 0.499 & 0.451 & 0.286 & 0.401 & \textcolor[rgb]{ 0,  .439,  .753}{\textbf{0.749}} \\
				\textbf{Prec(e$\leq$ 10)} & 0.52  & \textcolor[rgb]{ .329,  .51,  .208}{\textbf{0.829}} & 0.185 & \textcolor[rgb]{ 0,  .439,  .753}{\textbf{0.893}} & 0.771 & 0.663 & 0.602 & 0.71  & 0.489 & 0.603 & 0.542 & 0.559 & 0.464 & 0.51  & 0.462 & 0.299 & 0.416 & \textcolor[rgb]{ 1,  0,  0}{\textbf{0.894}} \\
				\textbf{Prec(e$\leq$ 20)} & 0.573 & \textcolor[rgb]{ .329,  .51,  .208}{\textbf{0.878}} & 0.366 & \textcolor[rgb]{ 0,  .439,  .753}{\textbf{0.901}} & 0.821 & 0.735 & 0.675 & 0.715 & 0.497 & 0.61  & 0.551 & 0.57  & 0.485 & 0.521 & 0.472 & 0.329 & 0.432 & \textcolor[rgb]{ 1,  0,  0}{\textbf{0.948}} \\
				\textbf{avg CP}  & 0.547 & \textcolor[rgb]{ .329,  .51,  .208}{\textbf{0.87}} & 0.722 & \textcolor[rgb]{ 0,  .439,  .753}{\textbf{0.889}} & 0.801 & 0.714 & 0.681 & 0.704 & 0.529 & 0.625 & 0.566 & 0.577 & 0.497 & 0.529 & 0.478 & 0.27  & 0.445 & \textcolor[rgb]{ 1,  0,  0}{\textbf{0.916}} \\
				\textbf{avg SR} & 0.57  & \textcolor[rgb]{ .329,  .51,  .208}{\textbf{0.871}} & 0.709 & \textcolor[rgb]{ 0,  .439,  .753}{\textbf{0.882}} & 0.807 & 0.772 & 0.765 & 0.703 & 0.65  & 0.627 & 0.596 & 0.577 & 0.562 & 0.542 & 0.496 & 0.489 & 0.447 & \textcolor[rgb]{ 1,  0,  0}{\textbf{0.923}} \\
				\hline
			\end{tabular}%
		}
		\caption{Results on POIC. Top-3 results of each dimension are colored in red, blue and green, respectively.}
		\label{tab:POIC_full_results}%
	\end{table*}%

	\subsection{Tracking}
	At the beginning of the tracking process, we crop the patch from the template containing the object, and then feed it into the network to obtain the feature $F_1(T)$ and $F_2(I)$ as described in Fig.~\ref{fig:all_structure} in the paper. The feature map is stored to save the computational cost since the highest computational burden comes from the feature extraction process.

	\begin{figure}[H]
		\centering
		\subfigure{
			\begin{minipage}[t]{0.46\linewidth}
				\centering
				\includegraphics[width=1\linewidth]{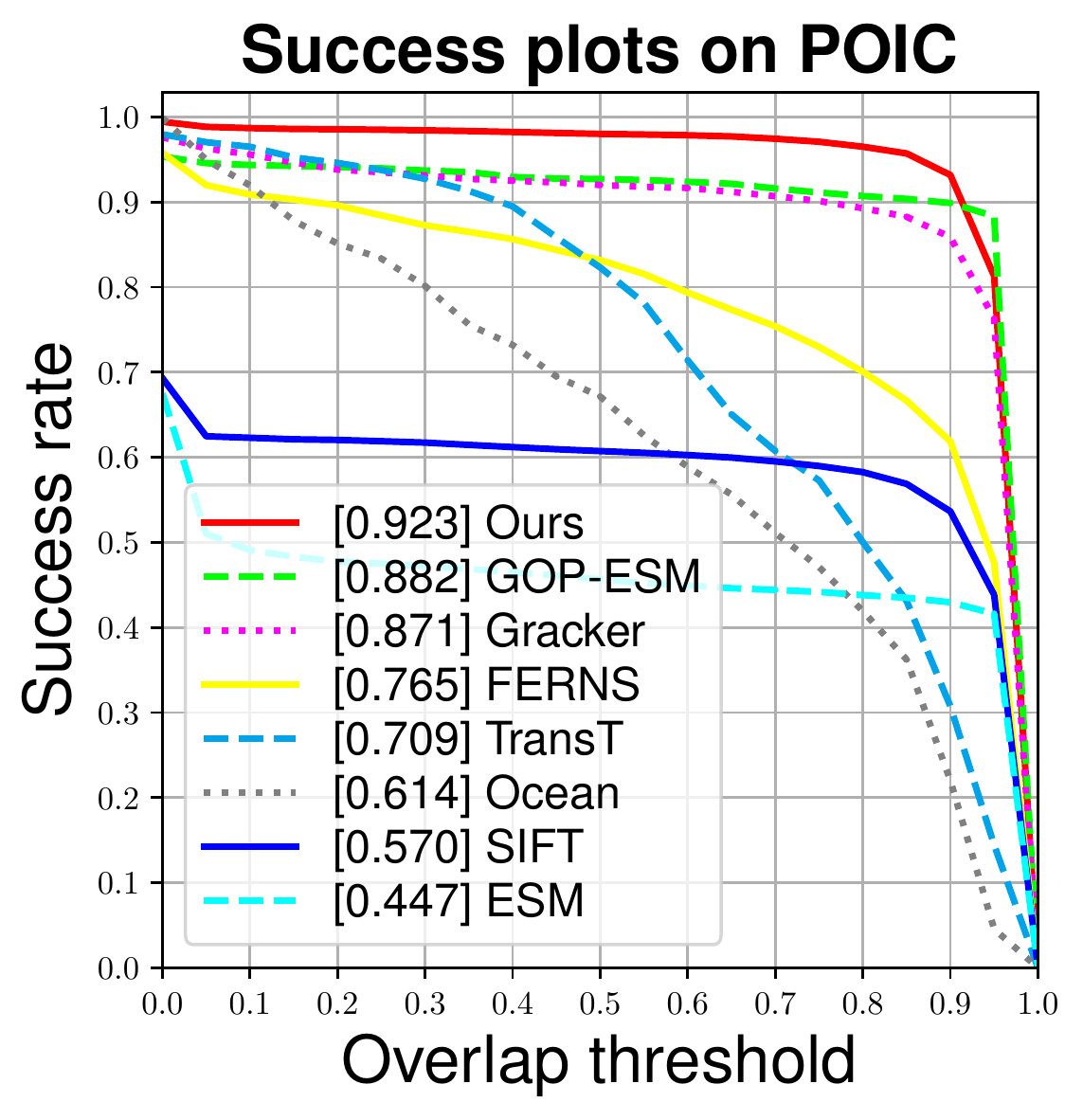}
				%\caption{fig1}
			\end{minipage}%
		}%
		\subfigure{
			\begin{minipage}[t]{0.46\linewidth}
				\centering
				\includegraphics[width=1\linewidth]{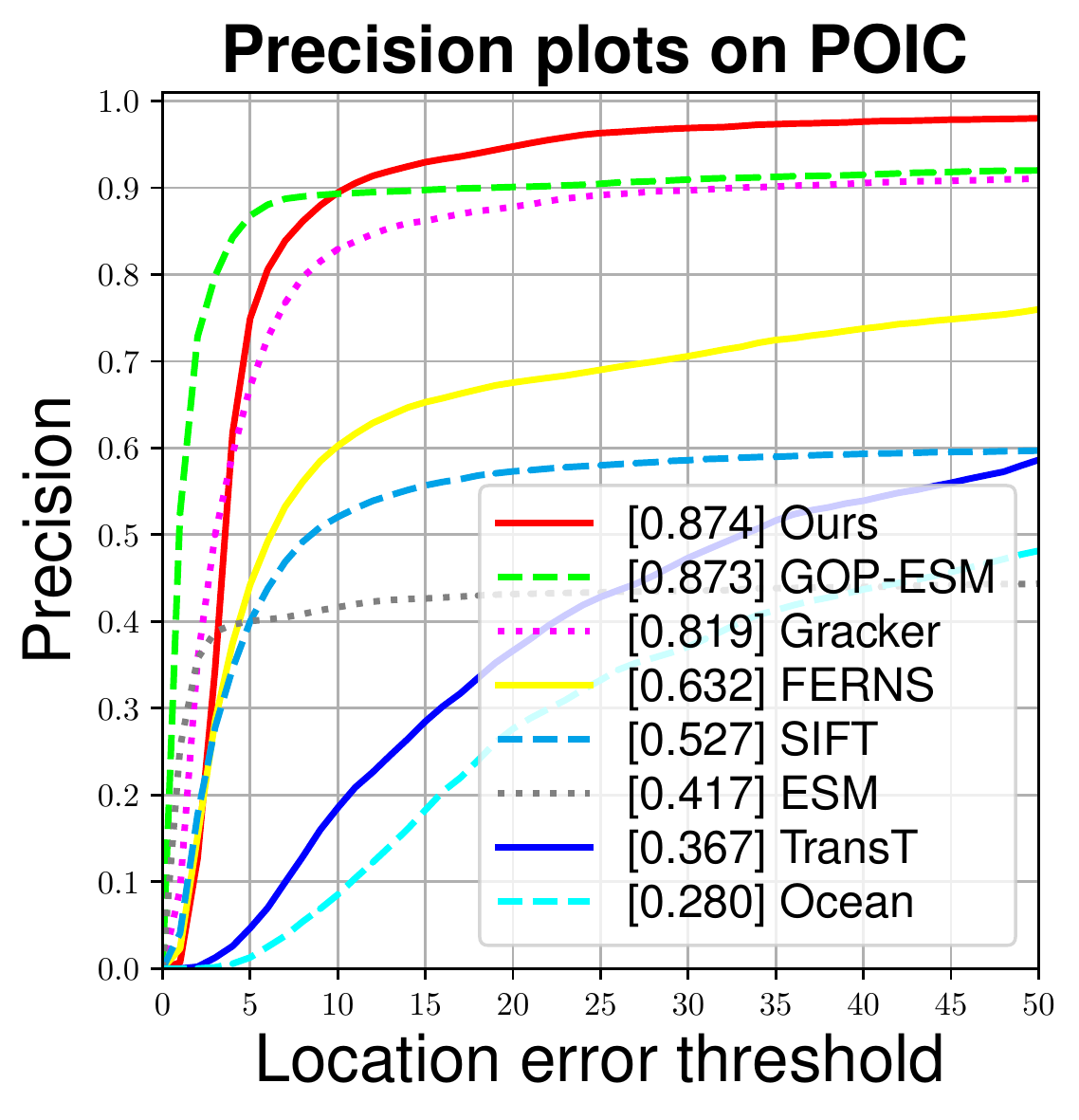}
				%\caption{fig2}
			\end{minipage}
		}%
		
		\centering
		\subfigure{
			\begin{minipage}[t]{0.46\linewidth}
				\centering
				\includegraphics[width=1\linewidth]{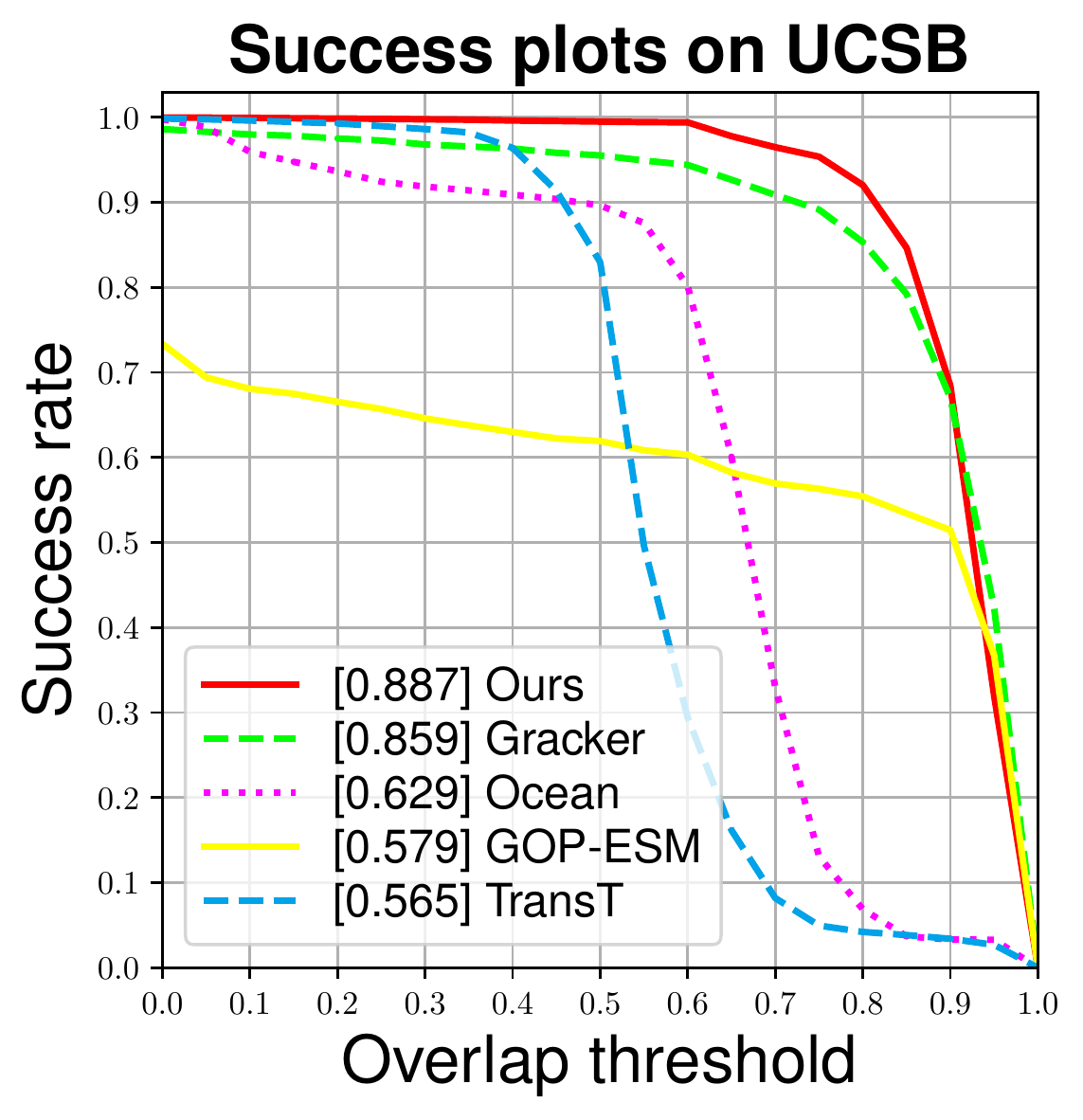}
				%\caption{fig2}
			\end{minipage}
		}
		\subfigure{
			\begin{minipage}[t]{0.46\linewidth}
				\centering
				\includegraphics[width=1\linewidth]{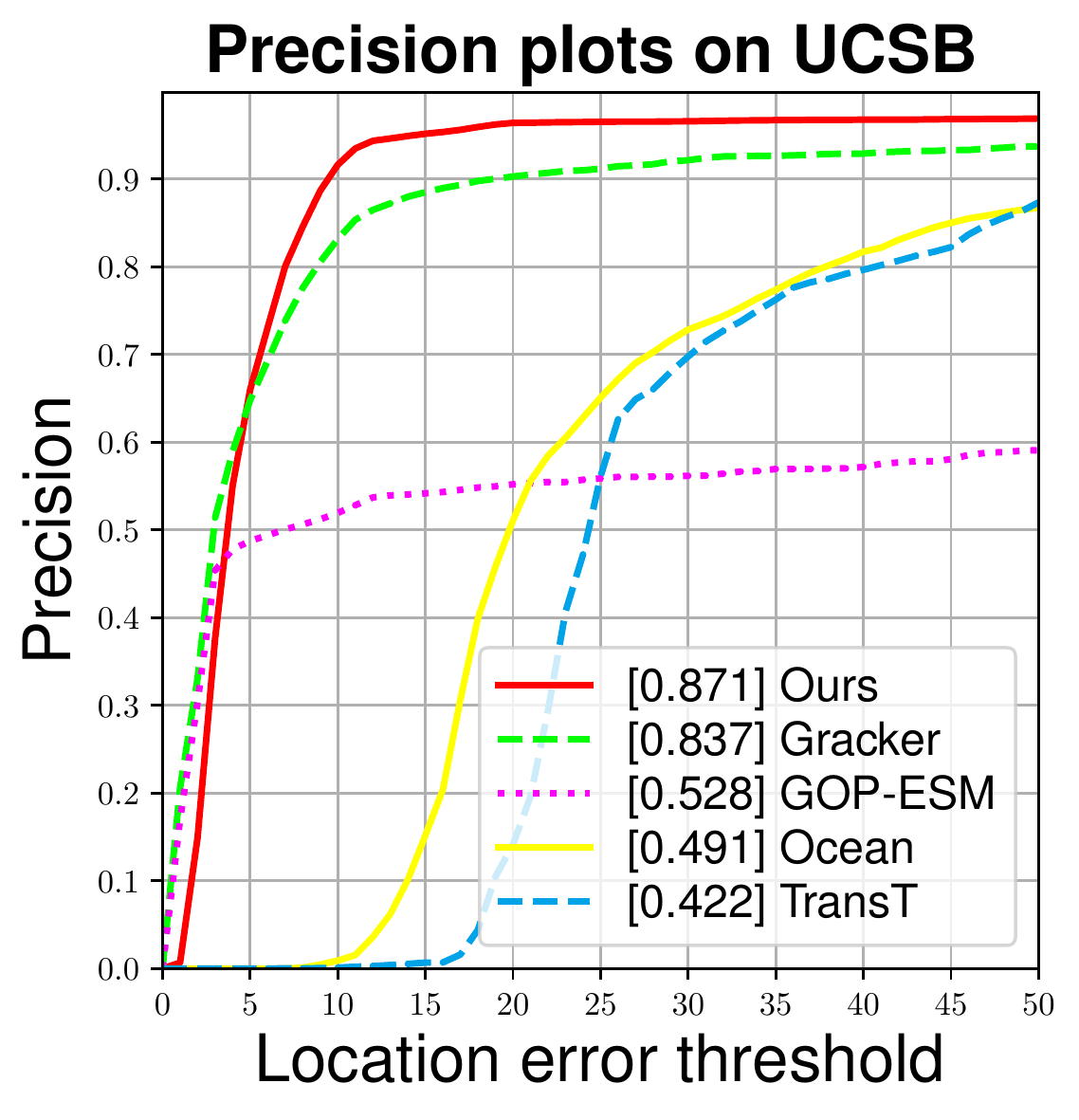}
				%\caption{fig2}
			\end{minipage}
		}%
		
		\centering
		\caption{SR and Prec on POIC and UCSB  datasets.}
		\label{fig:others_dat_compare}
% 		\vspace{-0.4in}
	\end{figure}
	In the subsequent frames, we firstly warp the image according to the current estimated homography $\mathbf{H}_i$, and crop the search patch and feed it to the translation branch.
	 Secondly, we estimate the object's center from the classification map and offset map, which is further used to warp the patch as input of the RSEW for generating the sampling grid and sampled image. Thirdly, the image is fed to accomplish scale and rotation estimation to obtain the $\mathbf{x}_S$, and then we warp the target according to this transformation. Finally, we estimate the real residual transformation between the warped image and template, where the compositional $\mathbf{H}$ could be calculated from Eq.~(1) in the main paper. 

	%	as the accumulated H for the next frame tracking. 
% 	\textcolor{blue}{
	\subsection{Evaluation}
	For the experiments on tracking precision and success rate, the precision is calculated under different corner alignment error thresholds, which is shown as the curve in Fig.~\ref{fig:pot_compare} in the main paper. For average precision, we take the average of all the precision with a different threshold. In the experiment on robustness, we intend to find out the different trajectory length ratios, thus a lower IoU $\leq0.2$ is adopted to denote a tracking failure. Note that we have tried different IoU thresholds, where the rankings are quite similar.
% 	}

	\section{Detailed Results}
	
	\subsection{Qualitative Results}
	Fig.~\ref{fig:results_plot_all} gives visual results compared with the state-of-the-art planar trackers. Our HDN method decomposes the homography and exploits the robust rotation and scale estimation by combining a deep homography estimator. Therefore, it obtains robust tracking results in contrast to the keypoint-based method~(SIFT, LISRD) and direct approach~(GOP-ESM).	
	
	To demonstrate the efficacy of our proposed approach, we replace the planar objects in POT with video advertisements. Please kindly check the attached videos (demo.mp4).
	
	\subsection{Quantitative Results}

	Fig.~\ref{fig:pot_compare_more} shows more scenarios of Precision and Homography Success rates on POT with increasing thresholds. Fig.~\ref{fig:others_dat_compare} gives the curves of SR and Prec on POIC and UCSB with increasing thresholds. These results are consistent with the analysis in the paper.
% 	\textcolor{blue}{

	Table.~\ref{tab:POT_full_results} shows the complete comparisons on POT in our experiment, and Table.~\ref{tab:POIC_full_results}
	reveals the complete comparisons on POIC~\cite{GOP-ESM}.
	Please check \cite{POT280} and \cite{GOP-ESM} for detailed trackers descriptions. Some trackers aren't included in the results for two reasons: 1) no source code available, e.g. CLKN~\cite{CLKN}. 2) As most of them consider the problem of image alignment, the tracking results are strongly related to the implementation of the tracking process, e.g. GEO, SuperGlue. We tried a straightforward implementation of SuperGlue~\cite{SuperGlue}. However, the result on POT is 0.464, which is lower than the result (0.552) reported in literature. Note that, most of these methods are not directly designed for planar object tracking.

	\section{Failure Cases}
	HDN is designed for planar object tracking with homography transformation, which may not be effective for general objects with large deformations. The typical failure happens when tracking planar objects with less texture, which usually don’t have enough visual features. Besides, very large occlusions lead to the estimation error on the object’s center. In this case, the regression results on the corners occluded are not reliable. Except that, if the object’s size in the first frame is too small, the error could also be large as a result of the low resolution and bad feature quality.
\end{document}